\documentclass[10pt,journal,compsoc]{IEEEtran}
\ifCLASSOPTIONcompsoc
  \usepackage[nocompress]{cite}
\else
  \usepackage{cite}
\fi

\ifCLASSINFOpdf
\else
\fi
\usepackage{amssymb}
\usepackage{amsmath}
\usepackage{algorithmic}
\usepackage{array}
\usepackage{stfloats}
\usepackage{url}
\usepackage{bm}
\usepackage{amsbsy}
\usepackage{multirow}
\usepackage{threeparttable}
\usepackage{dsfont}
\usepackage[switch]{lineno}
\usepackage{graphicx}
\usepackage[pagebackref=false,breaklinks=true,colorlinks,urlcolor=blue,citecolor=blue,linkcolor=blue,bookmarks=false]{hyperref}
\usepackage{makecell}

\begin{document}
%
\title{SignBERT+: Hand-model-aware Self-supervised Pre-training for Sign Language Understanding}
%
%
%
%

\author{Hezhen~Hu,
        Weichao~Zhao$^*$,
        Wengang~Zhou,~\IEEEmembership{Senior~Member,~IEEE,}
        and~Houqiang~Li,~\IEEEmembership{Fellow,~IEEE}
\IEEEcompsocitemizethanks{
\IEEEcompsocthanksitem This work was supported by the National Natural Science Foundation of China under Contract U20A20183 and 62021001. 
It was also supported by GPU cluster built by MCC Lab of Information Science and Technology Institution, USTC, and the Supercomputing Center of the USTC.

\IEEEcompsocthanksitem Hezhen Hu, Weichao Zhao, Wengang Zhou and Houqiang Li are with the Department of Electronic Engineering and Information Science
of Electrical and Computer Engineering, University of Science and Technology of China, Hefei, Anhui, 230026.\protect\\
E-mail: \{alexhu, saruka\}@mail.ustc.edu.cn, \{zhwg, lihq\}@ustc.edu.cn \\
$^*$Equal contribution with the first author. \\
Corresponding authors: Wengang Zhou and Houqiang Li.
}
}

%
%

\markboth{IEEE Transactions on Pattern Analysis and Machine Intelligence, April~2023}%
{Shell \MakeLowercase{\textit{et al.}}: Bare Demo of IEEEtran.cls for Computer Society Journals}
%



\IEEEtitleabstractindextext{%
\begin{abstract}
Hand gesture serves as a crucial role during the expression of sign language.
Current deep learning based methods for sign language understanding~(SLU) are prone to over-fitting due to insufficient sign data resource and suffer limited interpretability.
In this paper, we propose the \emph{first} self-supervised pre-trainable SignBERT+ framework with model-aware hand prior incorporated.
In our framework, the hand pose is regarded as a visual token, which is derived from an off-the-shelf detector.
Each visual token is embedded with gesture state and spatial-temporal position encoding.
To take full advantage of current sign data resource, we first perform self-supervised learning to model its statistics.
To this end, we design multi-level masked modeling strategies~(joint, frame and clip) to mimic common failure detection cases.
Jointly with these masked modeling strategies, we incorporate model-aware hand prior to better capture hierarchical context over the sequence.
After the pre-training, we carefully design simple yet effective prediction heads for downstream tasks.
To validate the effectiveness of our framework, we perform extensive experiments on three main SLU tasks, involving isolated and continuous sign language recognition~(SLR), and sign language translation~(SLT).
Experimental results demonstrate the effectiveness of our method, achieving new state-of-the-art performance with a notable gain.

\end{abstract}

}

\maketitle

\IEEEdisplaynontitleabstractindextext

%
\IEEEpeerreviewmaketitle

\IEEEraisesectionheading{\section{Introduction}\label{sec:introduction}}
\IEEEPARstart{S}{ign} language~(SL) serves as a primary communication tool for the deaf community.
It is a visual language with its unique grammar and lexicon, which is non-trivial for the hearing people to master.
To facilitate barrier-free communication between hearing and deaf people, automatic visual sign language understanding, as a topic with broad social influence, has been widely studied.
Visual sign language understanding contains three main tasks, including isolated and continuous sign language recognition~(SLR) and sign language translation~(SLT).
Isolated SLR focuses on word level recognition, which is essentially a fine-grained classification task.
Differently, continuous SLR targets at recognizing the sign gloss sequence with its corresponding occurring order, which needs to learn the sequence correspondence across the visual and textual domains.
SLT intends to further generate spoken language translations, which emphasizes natural linguistic expression.
These three tasks are all important for sign language understanding and bring challenges from different perspectives.

Hand gesture plays a dominant role during the meaning expression of sign language.
Intrinsically, hand occupies a relatively small spatial size, exhibiting uniform appearance and self-occlusion among joints.
During SL expression, it usually occurs over complex backgrounds and presents fast motion.
These characteristics lead to difficulty in hand representation learning.
Current deep-learning-based methods adaptively learn hand feature representations from the cropped RGB sequence.
Meanwhile, considering the highly-articulated characteristic of hand, some methods propose to utilize pose as the hand representation.
Pose is a compact and semantic representation, which is robust to appearance change and brings potential computation efficiency.
However, current pose-based methods utilize the poses extracted from off-the-shelf pose detectors, which usually suffer failure detection due to the motion blur and complex backgrounds, \emph{etc.}
Therefore, their performances usually lag largely behind the RGB-based counterparts.
Besides, the aforementioned methods all follow a data-driven paradigm and suffer over-fitting due to limited sign data resource and insufficient interpretability.

Meanwhile, the effectiveness of pre-training has been validated in computer vision~(CV) and natural language processing~(NLP) tasks.
Recent advances in NLP are largely derived from self-supervised pre-training techniques on large text corpus~\cite{radford2018improving,devlin2018bert,yang2019xlnet}.
Among them, BERT~\cite{devlin2018bert} is one of the most popular methods for its simplicity and effectiveness.
Its success is largely attributed to the strong Transformer backbone~\cite{vaswani2017attention} and well-designed pre-training strategies to model the context inherent in the text corpus.
By adding a simple head~(an MLP) on top for fine-tuning, it achieves notable performance gains in many downstream tasks, especially those with limited data resource. 
Notably, natural language is represented by a 1D sequence of text words which are characterised with well-defined semantic meaning. However, sign video is a kind of 3D data with complex spatial-temporal context, and it is non-trivial to analogically define visual word or unit with clarified semantics.
Therefore, it remains a hard challenge to leverage the success of BERT to video-based sign language understanding.

\begin{figure*}
	\centering
	\includegraphics[width=1.0\linewidth]{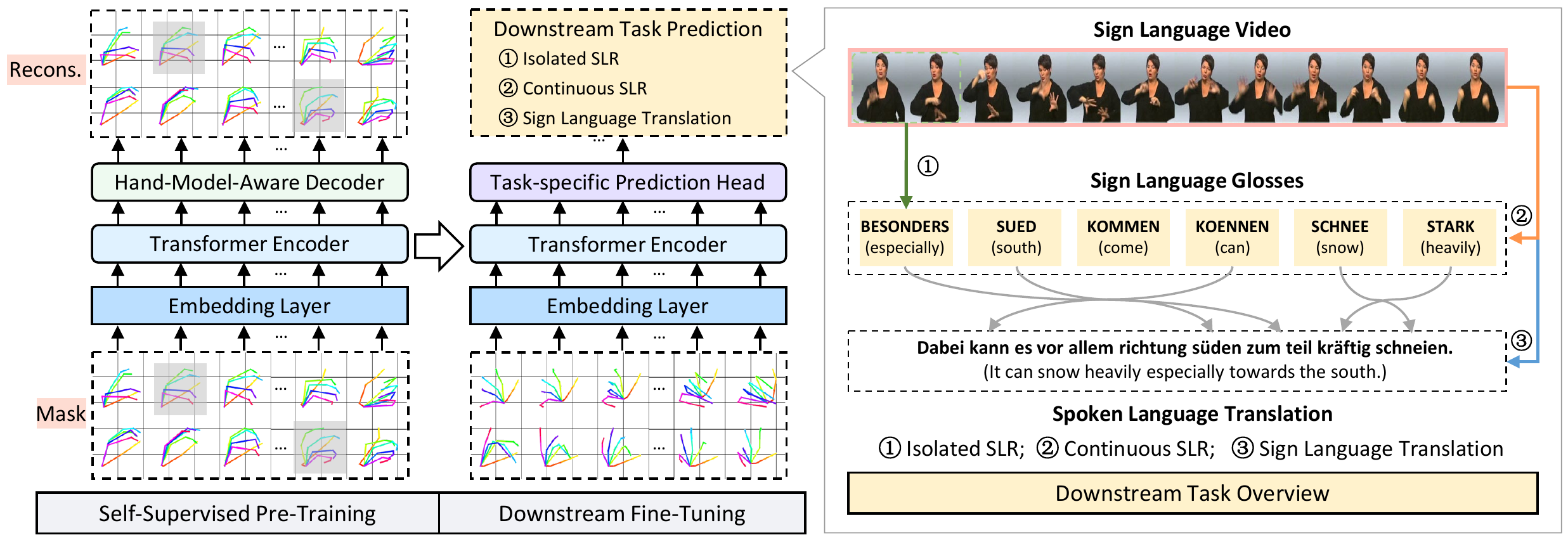}
	\caption{{The overview of our method and sign language understanding tasks~(isolated SLR, continuous SLR and SLT).}}
	\vspace{-1em}
	\label{fig:intro}
\end{figure*}

To tackle the above-mentioned issues, we develop a self-supervised pre-trainable framework with model-aware hand prior incorporated, namely SignBERT+, as shown in Figure~\ref{fig:intro}.
Considering the dominance of hand during SL expression, we utilize the compact and expressive hand pose as a visual token in a frame-by-frame manner.
Then, we carefully depict it with gesture state and spatial-temporal global position information.
SignBERT+ first performs self-supervised pre-training on a large volume of hand pose data, which is derived from an off-the-shelf detector on sign videos.
Inspired by the success of BERT~\cite{devlin2018bert}, we pre-train the encoder-decoder backbone via reconstructing the masked visual tokens from the corrupted input sequence, which enforces the framework to capture the hierarchical context in the sign language domain.
Considering the noisy characteristic of detected hand pose data, we carefully design multi-scale masking strategies, including joint, frame and clip levels.
Meanwhile, to better mine the context in the sign video domain, we further incorporate hand prior in a model-aware method.
After pre-training, we carefully design simple yet effective task-specific prediction heads, which are jointly fine-tuned with the pre-trained SignBERT+ encoder to adapt to downstream tasks.

In summary, our contributions are three-fold as follows,
\vspace{-0.5em}
\begin{itemize}
    \item To our best knowledge, we propose the \emph{first} model-aware pre-trainable framework, namely SignBERT+.
    It performs self-supervised pre-training on a large volume of sign pose data, followed by fine-tuning to achieve better performance on multiple downstream tasks.
    
    \item To better model the hierarchical context underneath the sign data during pre-training, we design multiple masked modeling strategies ranging from joint to clip level, in coordination with incorporated model-aware hand prior and spatial-temporal position encoding.
    For diverse downstream tasks, we design simple yet effective task-specific prediction heads on top of the pre-trained SignBERT+ encoder.
    
    \item We perform extensive experiments to validate the feasibility and effectiveness of our framework.
    Experimental results demonstrate that our method achieves new state-of-the-art performance on video-based sign language understanding tasks, including isolated SLR, continuous SLR and SLT.
\end{itemize}

This work is an extension of the conference paper~\cite{hu2021signbert} with improvement in a number of aspects.
1) Considering the characteristics of sign language, we further introduce spatial-temporal global position encoding into embedding, along with the masked clip modeling for modeling temporal dynamics.
Those new techniques further bring a notable performance gain.
2) We extend the original framework to two more downstream tasks in video-based sign language understanding, \emph{i.e.,} continuous SLR and SLT.
To this end, we design simple yet effective task-specific prediction heads.
Besides, we also provide efficient fusion strategies with the RGB modality.
Our newly designed framework achieves state-of-the-art performance on all the downstream tasks.
3) We present more comprehensive discussion on related works and make deep analysis on different components of our method to highlight the important ingredients.
Besides, we add discussions on future works and broader impact.

\section{Related Work}
In this section, we first give a literature review for video-based sign language understanding.
Then we present an overview of pre-training strategies.
Finally, we introduce related hand modeling techniques.

\subsection{Video-based Sign Language Understanding}
Video-based sign language understanding has made remarkable progress~\cite{koller2020quantitative,li2020transfer,min2021visual,camgoz2020sign}.
Generally, it contains three main tasks, including isolated SLR, continuous SLR and SLT.
These tasks emphasize different aspects, bringing their specific challenges to resolve.

\noindent \textbf{Isolated sign language recognition.}
Isolated SLR aims to recognize at the word level, which is essentially a fine-grained classification problem.
This task poses a challenge on learning discriminative visual representation~\cite{pfister2013large,lichtenauer2008sign,li2020transfer,bilge2022towards,albanie2020bsl}.
Early works utilize hand-crafted features, \emph{e.g.} HOG~\cite{koller2015continuous} and SIFT~\cite{pfister2013large}, to represent hand shape, orientation and motion.
Recently, researchers have resorted to deep learning techniques, which adaptively extract features from the full video sequence.
Based on the input modality, these works can be divided into RGB-based and pose-based methods.
RGB-based methods usually adopt Convolutional Neural Networks~(CNNs) as the backbone. 
{For instance, Koller~\emph{et al.}~\cite{koller2019weakly} utilize 2D-CNNs with LSTM to sequentially model the spatial and temporal representations.}
Some other works utilize 3D-CNNs for modeling spatial-temporal dependency~\cite{huang2018attention, joze2018ms, li2020transfer, li2020word, albanie2020bsl}.

For the pose-based counterpart, there exist different backbones for feature extraction, including CNNs~\cite{li2018co, albanie2020bsl} and RNNs~\cite{du2015hierarchical,min2020efficient,song2017end}, \emph{etc.}
Recently, considering its well-structured nature, more and more works have utilized graph convolutional networks~(GCNs), which exhibit both efficiency and effectiveness~\cite{du2015hierarchical, song2017end,tunga2020pose}.
As a representative work, ST-GCN~\cite{yan2018spatial} organizes the pose sequence as a pre-defined graph and adopts GCNs to perform recognition.
Besides, Tunga~\emph{et al.} further combines Transformer without pre-training for isolated SLR~\cite{tunga2020pose}.

\noindent \textbf{Continuous sign language recognition.}
It aims to map the sign video to the gloss sequence in the same presenting order.
In this task, the transitions between sign glosses may come with temporal variants, and the sign video usually lacks the frame-level gloss annotation.
Therefore, it raises a new challenge on the sequence correspondence learning between the visual sign representation to the sign glosses.
{To this end, Koller~\emph{et al.}~\cite{koller2017re,koller2018deep} exploit the integration of 2D-CNNs and Hidden Markov Models~(HMMs) for modeling transitions.}
Connectionist Temporal Classification~(CTC) is a differentiable cost function, which is able to deal with two unsegmented sequences without precise alignment.
It usually works with Recurrent Neural Networks~(RNNs), \emph{e.g.} BLSTM~\cite{hochreiter1997long} and GRU~\cite{cho2014properties}, and Transformer~\cite{vaswani2017attention} for sequential learning.
CTC-based methods make end-to-end optimization possible and become the mainstream for its competitive performance~\cite{cui2019deep, pu2020boosting, min2021visual,shi2019fingerspelling,camgoz2020sign}.
However, these methods are prone to over-fitting due to limited data resources.
To tackle this issue, DNF~\cite{cui2019deep} utilizes the iterative optimization strategy for better feature representation.
Zhou~\emph{et al.}~\cite{9635818} boosts the visual encoder with the partially masked videos under the supervised classification task.
CMA~\cite{pu2020boosting} proposes cross modality augmentation, which leverages the pseudo video-text pairs to boost recognition performance.
VAC~\cite{min2021visual} proposes visual alignment constraint to enhance the feature extractor.

\noindent \textbf{Sign language translation.}
This task intends to generate the spoken language translations.
It is mainly different from continuous SLR in the aspect of sequential learning, due to different grammar and word order between sign language and spoken language~\cite{cihan2018neural,li2020tspnet}.
NSLT~\cite{cihan2018neural} first explores this task with an attention-based encoder-decoder and proposes RWTH-PhoenixT dataset.
This dataset provides both sign gloss and translation annotation, and becomes the most popular benchmark.
Camgoz~\emph{et al.}~\cite{camgoz2020sign} leverage the strong modeling capability of Transformer into sequential learning.
TSPNet~\cite{li2020tspnet} explores the temporal semantic structures for more discriminative features.
STMC~\cite{zhou2021spatial} fuses information from multi-cue streams to boost performance.
SignBT~\cite{zhou2021improving} utilizes external text corpus for performance improvement.
In this work, we aim to leverage a large volume of sign data via pre-training to benefit three main sign language understanding tasks.

\subsection{Pre-Training Strategy}
Pre-training, as a common strategy in CV and NLP, aims to learn generic representation from massive labeled or unlabeled data, which benefits downstream tasks with marginal fine-tuning cost.
For fully supervised pre-training, it is common for CV tasks to first pre-train CNNs under labeled classification benchmarks, \emph{e.g.} ImageNet~\cite{deng2009imagenet} and Kinetics~\cite{carreira2017quo}, \emph{etc.}
However, given the labeling cost, more and more works turn to self-supervised learning from a large volume of unlabeled data, which is readily available from the Web~\cite{liu2021self, jing2020self}.
Self-supervised learning aims to model the joint probability distribution inherent in data, which is beneficial to address the following discriminative learning task. 

Pioneering works subtly design pretext tasks to perform self-supervised pre-training~\cite{zhang2016colorful,wang2021self,gidaris2018unsupervised,noroozi2016unsupervised, wang20193d,qi2020learning,misra2016shuffle}.
These tasks include predicting colorization~\cite{zhang2016colorful}, rotation ~\cite{gidaris2018unsupervised}, transformation~\cite{qi2020learning} and frame / clip orders~\cite{misra2016shuffle, xu2019self}, \emph{etc.}
Recently, some works focus on contrastive learning for pre-training~\cite{he2020momentum,li2021crossclr,chen2021exploring}.
Typically, it aims to pull the representation of similar instances closer, while pushing away negative instances.
To obtain informative negative instances for better optimization, some works utilize the techniques like memory banks~\cite{he2020momentum} and large batch size~\cite{chen2020simple}.
There also exist works further eliminating the requirement of negative samples~\cite{grill2020bootstrap, chen2021exploring}.

Another interesting strand is the generative self-supervised pre-training, which usually involves training the encoder via the reconstruction task.
In NLP, one milestone of pre-training is BERT~\cite{devlin2018bert}.
BERT is built on the strong Transformer backbone with masked language modeling~(MLM) as one of its pre-training tasks.
MLM attempts to predict the masked words by leveraging the context cues from the remaining tokens.
Similar to BERT, some works also adopt MLM during pre-training, such as GPT~\cite{radford2018improving}, XLNet~\cite{yang2019xlnet} and RoBERTa~\cite{liu2019roberta}, \emph{etc.}
These pre-training methods generalize well and bring notable performance gains on the downstream tasks.
Motivated by the success in NLP, some works attempt to leverage the idea of BERT into CV tasks~\cite{sun2019videobert,Su2020VLBERT,chen2020generative,dosovitskiy2020image, bao2021beit,he2021masked}, which mainly focus on the RGB modality.
BEiT~\cite{bao2021beit} utilizes the discrete tokenized image patches as pseudo labels and performs masked modeling similar to BERT.
MAE~\cite{he2021masked} directly works in the continuous space, \emph{i.e.,} masking and reconstructing the pixel values.
However, BEiT and MAE only focus on image-based tasks.
Actually, it is non-trivial to leverage BERT's success to video-based sign language understanding tasks, which involves special design of the pretext task and framework architecture.

\noindent \textbf{Pre-training in sign language.}
Albanie~\emph{et al.}~\cite{albanie2020bsl} propose to perform supervised pre-training on a large-scale annotated dataset.
Li~\emph{et al.}~\cite{li2020transfer} boost isolated SLR via a domain-invariant feature descriptor, which leverages the knowledge from external subtitled news sign video.
To our best knowledge, there exists no self-supervised pre-training works in the sign language domain.

\subsection{Hand Modeling Techniques}
Hand modeling aims to depict the hand with more expressiveness.
Current modeling techniques include sum-of-Gaussians~\cite{sridhar2013interactive}, shape primitives~\cite{oikonomidis2014evolutionary, qian2014realtime} and sphere-meshes~\cite{tkach2016sphere}, \emph{etc.}
To better reconstruct the hand shape, Iason~\emph{et al.}~\cite{oikonomidis2011efficient} define scaling terms on bone lengths.
Notably, some works~\cite{ballan2012motion, tzionas2016capturing, romero2017embodied} attempt to learn hand shape variation with Linear Blend Skinning~(LBS)~\cite{lewis2000pose}.
Among them, MANO~\cite{romero2017embodied} becomes the most popular one for its wide applications~\cite{habermann2020deepcap, boukhayma20193d,hu2021hand,moon2020interhand2}.
As a statistical model, it learns from a large variety of high-quality hand scans and represents the geometric changes in the low-dimensional pose and shape space.
With this capability, we adopt it as a constraint in the decoder to incorporate prior.

\section{Our Approach}
\begin{figure*}
	\centering
	\includegraphics[width=\linewidth]{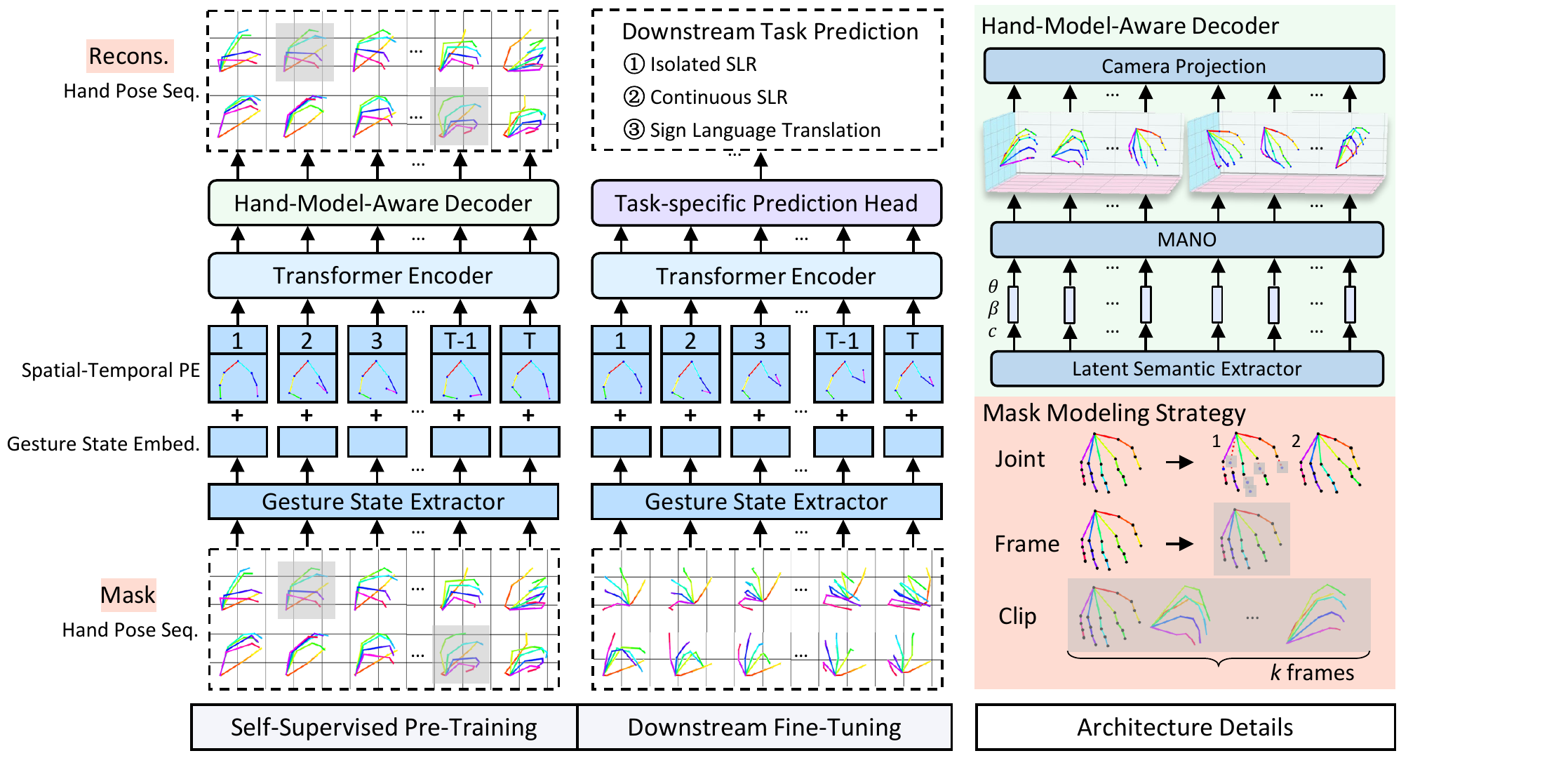}
	\caption{{Illustration of our SignBERT+ framework details, which contains self-supervised pre-training and fine-tuning for the downstream tasks.
  We organize the pre-extracted 2D poses of both hands as the visual token sequence.
For each token, it is embedded with gesture state and spatial-temporal position encoding.
During self-supervised pre-training, multi-level masked modeling strategies work with incorporated model-aware hand prior, in order to better capture the hierarchical context in the sign domain.
Given the downstream-task diversity, we design the task-specific prediction head and fine-tune it with the pre-trained SignBERT+ encoder.}}
	\label{fig:framework}
\end{figure*}

As shown in Figure~\ref{fig:framework}, our framework contains two stages, \emph{i.e.,} self-supervised pre-training and downstream task fine-tuning.
During sign expression, both hands are involved to act as a dominant role.
Therefore, we focus on them to build the visual token in a frame-wise manner.
For each visual token, we embed the gesture state and global spatial-temporal position information.
During pre-training, the whole framework works in a self-supervised paradigm by reconstructing the masked visual tokens from the corrupted input sequence.
Jointly with the multi-level masking strategies, we incorporate hand prior to better capture hierarchical context in the sign domain.
Then, we fine-tune the pre-trained SignBERT+ encoder~(embedding layer and Transformer encoder) with the designed prediction heads for downstream tasks.

In the following, we first introduce the framework architecture.
After that, we elaborate our pre-training strategy. 
Finally, we discuss the fine-tuning schemes for downstream tasks.

\subsection{Framework Architecture}
Our framework contains four main components, \emph{i.e.,} input embedding layer, Transformer encoder, hand-model-aware decoder and prediction head.

\subsubsection{Input Embedding Layer}
Given the dominant role of hand during sign language, we carefully design an embedding layer to capture cues from both hands.
It extracts gesture state and spatial-temporal position information from the hand pose sequence in a frame-wise manner, which are elaborated in the follow paragraphs.

\textbf{Gesture state embedding.}
Given its well-structured characteristics, we first organize the input 2D hand pose $\widetilde{J}_{t}$ at frame $t$ as an undirected spatial graph.
This graph is constructed with the node $V$ and edge $E$ set.
The node set includes all hand joints, \emph{i.e.,} 21 joints per hand, while the edge set contains their physical and symmetrical connection.
Then, the hand pose sequence is processed by a spectral-based GCN~\cite{cai2019exploiting, yan2018spatial}, which hierarchically performs graph convolution and graph pooling for gesture state embedding.
The graph convolution is formulated as follows,
\begin{equation}
\label{equ:gcn}
\bm{f}_{out}=\sum\limits_{i}{\mathbf{D}_{i}^{-\frac{1}{2}}\mathbf{{A}}_{i}\mathbf{D}_{i}^{\frac{1}{2}}\bm{f}_{in}\mathbf{{W}}_{i}},
\end{equation}
where $\bm{f}_{in}$ and $\bm{f}_{out}$ are the corresponding input and output features, respectively.
$i$ indicates the type of neighbors for each node.
$\mathbf{{W}}_{i}$ denotes the convolution weight and $\mathbf{A}_i$ is the dismantled matrix indicating the edge connection.
For graph pooling, we first cluster the original 21 joint nodes of each hand into 6 subsets corresponding to 5 fingers and 1 palm.
Then we perform max-pooling on the nodes in each subset, leading to 6 nodes.
Finally, these nodes are again max-pooled into one and both hands are involved in the frame-level gesture state embedding $\bm{f}_{p,t}$.

\textbf{Spatial-temporal position encoding.}
Besides the gesture state, hand spatial trajectory and temporal information also matter in video-based sign language understanding.
{We depict the hand global position in the normalized 2D space by introducing the arm joints of both sides.
These joints are also processed by GCN~\cite{cai2019exploiting, yan2018spatial} to extract the frame-level spatial embedding $\bm{f}_{s,t}$.}
Since the Transformer layers process the sequence in an order-agnostic way, we add temporal information into the input embedding $\bm{f}_{e,t}$, which is implemented by the position encoding technique in~\cite{vaswani2017attention}.

\subsubsection{Transformer Encoder}
The embedded input sequence is fed into the Transformer encoder~\cite{vaswani2017attention} for the latent semantic representation.
Its basic layer mainly contains two components, \emph{i.e.,} a multi-head self-attention module and a feed-forward network.
For each layer, its output retains the same size with the input.
The whole encoder is formulated as follows,
\begin{equation}
\begin{split}
	\mathbf{F}_0 &= \{\bm{f}_{p,t}+\bm{f}_{s,t}+\bm{f}_{e,t}\}_{t=1}^{T}, \\
	\widetilde{\mathbf{F}}_i &= L(M(\mathbf{F}_{i-1}) + \mathbf{F}_{i-1}), \\
	\mathbf{F}_i &= L(C(\widetilde{\mathbf{F}}_i) + \widetilde{\mathbf{F}}_i),
\end{split}
\end{equation}
where $i$ denotes the $i$-th layer of the Transformer encoder.
The whole encoder contains totally $N$ layers.
$M(\cdot)$, $C(\cdot)$ and $L(\cdot)$ represent the multi-head self-attention, feed-forward network and layer normalization, respectively.
$\mathbf{F}_i$ denotes the output feature from the $i$-th layer.

\subsubsection{Hand-model-aware Decoder}
To achieve the reconstruction target during pre-training, the decoder transforms the latent feature back to the pose sequence.
The decoder works in a model-aware method to incorporate prior, which aims to guide the encoder better capturing generic representations in the sign language domain.
Specifically, the latent feature is first processed by a fully-connected layer, which extracts the low-dimensional semantic embeddings depicting the hand status, \emph{i.e.,} hand pose $\bm{\theta}$ and shape $\bm{\beta}$, and the camera parameter $\bm{c}$ aligning the image plane, which is formulated as follows,
\begin{equation}
\label{equ:cnn_tcn}
  \mathbf{F}_{la} = \{\bm{\theta}, \bm{\beta}, \bm{c}_r, \bm{c}_o, \bm{c}_s\}_{t=1}^T = D(\mathbf{F}_N),
\end{equation} 
where $D(\cdot)$ denotes the fully-connected layer.
$\bm{\theta}$ and $\bm{\beta} \in \mathbb{R}^{10}$ are the hand pose and shape embeddings for the following MANO model, respectively.
$\bm{c}_r \in \mathbb{R}^{3\times3}$, $\bm{c}_o \in \mathbb{R}^{2}$, and $\bm{c}_s \in \mathbb{R}$ are parameters of the weak-perspective camera, representing the rotation, translation and scale, respectively.

Then the MANO model~\cite{romero2017embodied} incorporates hand prior and decodes the estimated hand embedding.
Specifically, the decoding process is fully-differentiable, which transforms the hand embedding~($\bm{\theta}$ and $\bm{\beta}$) to the dense triangular hand mesh $\mathbf{M} \in \mathbb{R}^{N_v \times 3}$~($N_v=$778 vertices and $N_f=$1538 faces) as follows,
\begin{equation}
\label{equ:mano}
  \mathbf{M}(\bm{\beta}, \bm{\theta}) = W(\mathbf{T}(\bm{\beta}, \bm{\theta}), J(\bm{\beta}), \bm{\theta}, \mathbf{W}),
  \vspace{-0.5cm}
\end{equation}
\begin{equation}
\label{equ:mano2}
  \mathbf{T}(\bm{\beta}, \bm{\theta}) = \bar{\mathbf{T}} + B_S(\bm{\beta}) + B_P(\bm{\theta}),
\end{equation}
where $B_S(\cdot)$ and $B_P(\cdot)$ represent shape and pose blend functions, respectively. 
$\mathbf{W}$ is a set of blend weights. 
The hand template $\bar{\mathbf{T}}$ is first posed and skinned based on the pose and shape corrective blend shapes, \emph{i.e.,} $B_P(\bm{\theta})$ and $B_S(\bm{\beta})$.
Then the mesh $\mathbf{M}$ is generated by rotating each part around joints $J(\bm{\beta})$ based on the linear skinning function $W(\cdot)$~\cite{kavan2005spherical}.
Besides, the sparse joint representation $\widetilde{J}_{3D}$ is also derived from the mesh.
To keep consistent with the input pose format, we further add 5 extra fingertip joints by selecting vertices with the index of 333, 443, 555, 678 and 734.
Finally, $\widetilde{J}_{3D}$ is mapped back to the same 2D plane as the input pose based on the estimated camera parameter as follows,
\begin{equation}
\label{equ:weak}
  \widetilde{J}_{2D} = \bm{c}_s\prod{({\mathbf{c}_{r}}{{\widetilde{J}}_{3D}})}+\mathbf{c}_o,
\end{equation} 
where $\prod(\cdot)$ denotes the orthographic projection.

\subsubsection{Prediction Head}
\label{sec: pred.head}
Given the large diversities among downstream tasks, we design simple yet effective prediction heads for each task in Figure~\ref{fig:PHead}.
In Section~\ref{sec: fine-tuning}, we will introduce them in detail along with the task-specific fine-tuning settings.

\subsection{Pre-Training SignBERT+}
In this section, we elaborate our pre-training paradigm.
Pre-training is performed via reconstructing the masked visual tokens from the corrupted input sequence, which aims to exploit hierarchical context on a large volume of sign pose data.
Different from the original BERT working on discrete word space, we attempt to pre-train on continuous pose space.
Therefore, it raises new issues to resolve, including the design of the masking strategies and objective functions.

\subsubsection{Masking Strategy}
Considering the noise of the detected input hand pose, the masking strategy needs to be carefully redesigned.
Given the hand pose sequence, we first randomly choose a portion $R$ of all tokens.
For the chosen token, one of the following operations is applied with the equal probability, \emph{i.e.,} masked joint modeling, masked frame modeling, masked clip modeling and identity modeling.

\textbf{Masked joint modeling.}
This strategy mimics the failure cases of pose detectors on some joints.
For a chosen token, we randomly choose $m$ joints.
Two operations are performed on these chosen joints with the equal probability, \emph{i.e.,} zero masking~(masking the coordinates of joints with zeros) or random spatial disturbance.
This modeling aims to guide the framework to infer the gesture state from the remaining joints, thus capturing the context at the joint level.

\textbf{Masked frame modeling.}
It aims to deal with the failure case on the whole frame pose, which is often caused by complex backgrounds.
The chosen token is directly zero masked.
This strategy enforces the framework to reconstruct the token by leveraging the observation from the remaining frames and the other hand. 
In this way, the temporal context for each hand and mutual context between hands are captured.

\textbf{Masked clip modeling.}
Motion blur, as a factor not to be overlooked, usually causes pose detection failure on a video clip.
To deal with this situation, masked clip modeling is designed.
We randomly choose $k$ temporally continuous tokens, where $k$ ranges from 2 to $K$.
The chosen $k$ tokens are all zero-masked.
In order to reconstruct them, the framework needs to capture the temporal dynamics by leveraging the motion pattern of existing frames.

\textbf{Identity modeling.}
Similar to BERT~\cite{devlin2018bert}, identity modeling directly feeds the unchanged tokens into the framework.
It is indispensable for the framework to learn identity semantic encoding on those unmasked tokens.

\subsubsection{Objective Functions}
During pre-training, its objective is to maximize the likelihood of the joint probability distribution to reconstruct the hand pose sequence.
To achieve the reconstruction target, the classification objective in the original BERT is substantially changed into regression.
To this end, we design the objective function as follows,
\begin{equation}
\label{equ:pre-train}
  \mathcal{L} = \mathcal{L}_{rec} + \lambda \mathcal{L}_{reg},
\end{equation}
where $\mathcal{L}_{rec}$ and $\mathcal{L}_{reg}$ are reconstruction and regularization loss terms, respectively.
$\lambda$ denotes the weighting factor.
We only include the corresponding output of the masked tokens during the loss calculation.

\textbf{Reconstruction loss $\mathcal{L}_{rec}$.}
Since the utilized pose usually contains noise due to failure detection, we utilize the detection confidence score as the filter to eliminate these influences.
The reconstruction loss is calculated as follows,
\begin{equation}
\small
\label{equ:rec}
  \mathcal{L}_{rec} = \sum\limits_{t, j}\mathds{1}(s(t,j)>=\epsilon)s(t,j){{\left\|\widetilde{J}_{2D}(t,j) - J_{2D}(t,j) \right\|}_{1}},
\end{equation}
where $\mathds{1}(\cdot)$ denotes the indicator function, and $s(t,j)$ denotes the detection confidence score of the ${J}_{2D}$ with joint $j$ at time $t$.
The joints with the confidence lower than $\epsilon$ are not included in the loss calculation.

\textbf{Regularization loss $\mathcal{L}_{reg}$.}
To ensure this decoder working properly, the regularization loss is added.
It is implemented by constraining the magnitude and derivative of the MANO input, which is responsible for generating the plausible mesh and keeping the signer identity unchanged.
This loss term is computed as follows,
\begin{equation}
\label{equ:reg}
  \mathcal{L}_{reg} = \sum\limits_{t}( {\left\|\bm{\theta}_t\right\|}_{2}^2 + w_{{\beta}}{\left\|\bm{\beta}_t \right\|}_{2}^2 + w_{{\delta}}{\left\|\bm{\beta}_{t} - \bm{\beta}_{t-1} \right\|}_{2}^{2}),
\end{equation}
where $w_{\beta}$ and $w_{\delta}$ denote the weighting factors.

\subsection{Fine-Tuning SignBERT+}
\label{sec: fine-tuning}
\begin{figure}
	\centering
	\includegraphics[width=1.0\linewidth]{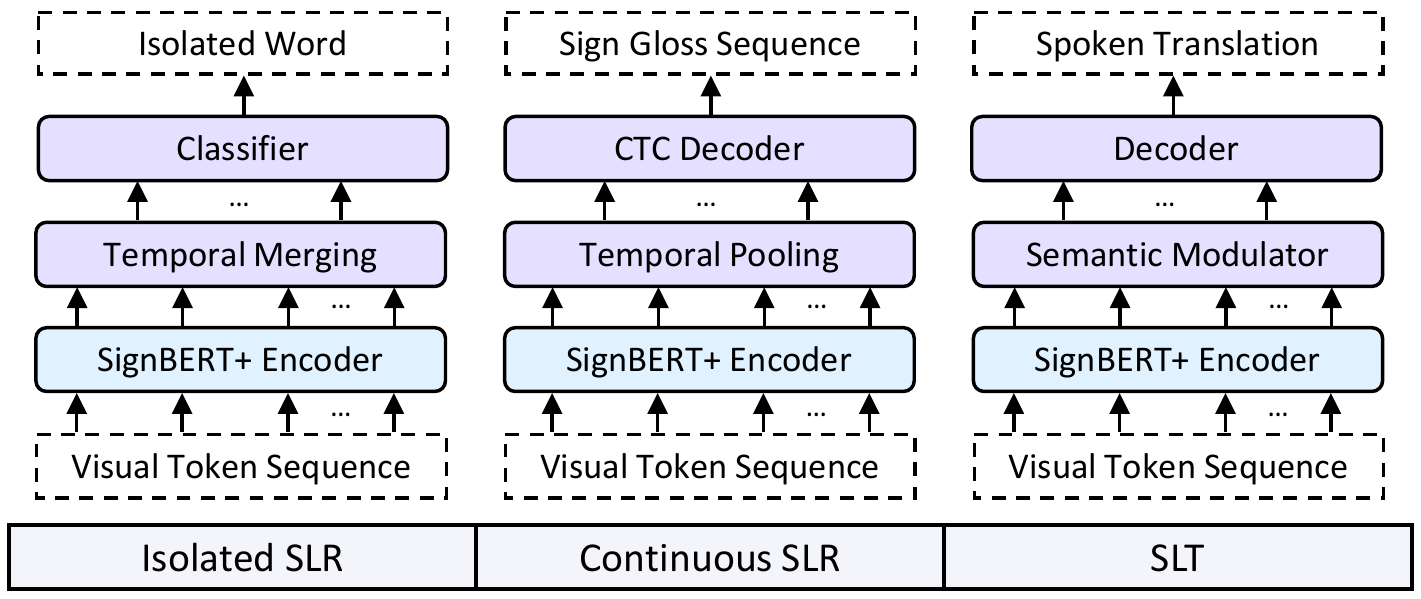}
	\vspace{-0.2cm}
	\caption{Illustration of the settings on three downstream tasks, \emph{i.e.,} isolated SLR, continuous SLR and SLT.
  The box in purple denotes our designed task-specific prediction head.
  It is fine-tuned with the pre-trained SignBERT+ encoder.
	}
	\label{fig:PHead}
\end{figure}

After pre-training SignBERT+ for generic visual representation in sign language, it is relatively simple to fine-tune it for various downstream tasks.
During fine-tuning, the task-specific prediction head is added on top of the pre-trained SignBERT+ encoder, as illustrated in Figure~\ref{fig:PHead}.
The framework is supervised under the task-specific loss.

Since only the hand pose modality is insufficient to convey the full meaning of sign language, we further provide the task-specific fusion strategy with the method based on the full RGB frames.
For clarity, the baseline RGB method utilized for fusion will be marked in the experiment section.
Besides, we denote our vanilla and fused framework as \textbf{Ours} and \textbf{Ours~(+ R)}, respectively.

\subsubsection{Isolated Sign Language Recognition}
Isolated SLR is a fine-grained classification problem, which categorizes a sign video to the corresponding isolated word.
For this task, the designed prediction head consists of a temporal merging module and a classifier.
The former module utilizes a simple attention mechanism to highlight the discriminative cues in certain frames during the merging process as follows,
\begin{equation}\label{equ:ISLR1}
  \bm{o} = Softmax(\mathbf{F}\mathbf{W} + b) \cdot \mathbf{F},
\end{equation}
where $\mathbf{F}$ and $\bm{o}$ denote the input feature sequence and merged feature, respectively.
Then the merged feature $\bm{o}$ is passed through a classifier~(MLP and softmax layer) to output the probability matrix.
Since isolated SLR is a classification problem, we utilize the cross-entropy loss to supervise the fine-tuning process.
We use the simple late fusion strategy with the RGB method, which directly sums their prediction scores and chooses the class with the highest score as the final recognition result.

\subsubsection{Continuous Sign Language Recognition}
Continuous SLR aims to recognize the gloss sequence $\bm{g}$ in the same presenting order as the sign actions in the input sign video $\mathbf{V}$ with $T$ frames.
The prediction head for this task contains a temporal pooling module and a CTC decoder.
The temporal pooling module aggregates frame-level visual features to the clip level, which outputs the one-quarter temporal length of the input.
Then it is fed into the connectionist temporal classification~(CTC) decoder to deal with the mapping between two unsegmented sequences without explicit alignment.

The objective of CTC is to maximize the posterior probability over all alignments from the source to the target.
It extends the vocabulary with a blank label to cover the cases of transition and silence.
Denote each alignment path of the input sequence as $\bm{\pi} = \{ {\bm{\pi}_t|}_{t=1}^T \}$ with $T$ as temporal duration.
Under the time independence assumption, its probability is computed as follows,
\begin{equation}\label{equ:ctc_path}
  p(\bm{\pi}|\mathbf{V})=\prod_{t=1}^{T} p(\bm{\pi}_t|\mathbf{V}).
\end{equation}
Typically, there exists many-to-one mapping from multiple input sequences to one target, which is achieved by removing all blanks and repetition.
In this way, we calculate the conditional probability of the target gloss sequence $\bm{g}$ by summing the probabilities of all possible mapping paths as follows,
\begin{equation}\label{equ:ctc_prob}
  p(\bm{g}|\mathbf{V}) = \sum_{\bm{\pi}\in \mathcal{B}^{-1}(\bm{s})} p(\bm{\pi}|\mathbf{V}),
\end{equation}
where $\mathcal{B}(\cdot)$ denotes the many-to-one mapping function.
$\mathcal{B}^{-1}(\cdot)$ is the inverse mapping of $\mathcal{B}(\cdot)$.
During training, the objective is defined by the negative log probability of $p(\bm{g}|\mathbf{V})$ as follows,
\begin{equation}\label{equ:ctc_loss}
  \mathcal{L}_{\mathrm{CSLR}} = -\ln p(\bm{g}|\mathbf{V}).
\end{equation}
During inference, the CTC decoder obtains a series of sentences via beam search and chooses the one with the highest decoding probability as the final prediction.

For fusion, similar to~\cite{camgoz2017subunets}, we first concatenate the encoded feature from the RGB baseline and our method.
Then we utilize a BLSTM sequential model and a CTC decoder to map the merged feature to the gloss sequence.

\subsubsection{Sign Language Translation}
Given the input sign video $\mathbf{V}$ with $T$ frames, SLT aims to generate the spoken language translation $\bm{s}=\{\bm{s}_i\}_{i=1}^N$ with $N$ words via maximizing the conditional probability $p(\bm{s}|\mathbf{V})$.
For this task, our designed prediction head contains a semantic modulator and a decoder as shown in Figure~\ref{fig:PHead}.

{Considering the token length diversity between the source and target~($T >> N$), the semantic modulator attempts to bridge this gap and generates suitable semantics $\mathbf{M}=\{\bm{m}_i\}_{i=1}^{T_1}$ for the decoder.
Specifically, it first performs average temporal pooling to reduce the source visual token sequence from the length $T$ to the length $T_1=T/4$.
This operation makes the visual representation more compact, but its output lacks temporal dependency modeling.
To mitigate this issue, a Transformer encoder is adapted to further modulate the pooled visual sequence and generate suitable semantics.}

{After that, a decoder is adopted to perform mapping between sign language and spoken translation while considering their different grammar.
The decoder contains two main components, \emph{i.e,} a word embedding layer and an autoregressive Transformer decoder.
The word embedding layer embeds each word in the target sequence $\bm{s}$, along with the added position encoding as follows,
\begin{equation}\label{equ:dec_embed}
  \bm{w_i} = WordEmbed(\bm{s}_i) + PE(i),
\end{equation}
where $\bm{s}_i$ denotes the input word, $WordEmbed(\cdot)$ and $PE(\cdot)$ are the word embedding and position encoding functions, respectively.}

{The autoregressive property means the model leverages generated text as additional input when generating the next.
The Transformer decoder architecture is also a stack of basic blocks.
The basic block contains three components, \emph{i.e.,} masked multi-head self-attention module, multi-head cross-attention module and feed-forward network.
The mask adopted on self-attention ensures the information flow in the rightward direction to preserve the autoregressive property~\cite{vaswani2017attention,graves2013generating}.
This operation is necessary for the SLT inference, since the framework is not accessible to the output tokens which are decoded currently or in the future.
Cross-attention module leverages the contextual cues from the modulated visual semantics $\mathbf{M}$ and predecessors words $\bm{w}$.
The whole Transformer decoder is formulated as follows,
\begin{equation}
\begin{split}
	\mathbf{D}_0 &=  \bm{w}, \\
	{\mathbf{Q}}_i &= L(\widetilde{M}(\mathbf{D}_{i-1}) + \mathbf{D}_{i-1}), \\
	\widetilde{\mathbf{D}}_i &= L(MHA({\mathbf{Q}}_i, \mathbf{M}^k, \mathbf{M}^v) + {\mathbf{Q}}_i), \\
	\mathbf{D}_i &= L(C(\widetilde{\mathbf{D}}_i) + \widetilde{\mathbf{D}}_i),
\end{split}
\end{equation}
where $i$ denotes the $i$-th layer of the Transformer decoder.
The whole encoder contains totally $N$ layers.
$\widetilde{M}(\cdot)$, $MHA(\cdot)$, $C(\cdot)$ and $L(\cdot)$ represent the masked multi-head self-attention, multi-head cross-attention, feed-forward network and layer normalization, respectively.
$\mathbf{D}_i$ denotes the output feature from the $i$-th layer.}

During decoding, the sentence is first prefixed with the word ``[bos]'' to indicate the beginning.
Then each word in the target sequence $\bm{s}$ is embedded.
The embedded sequence is then fed into the Transformer decoder $TransD(\cdot)$.
This decoder additionally performs cross attention by leveraging the contextual cues from the modulated visual semantics $\bm{m}$ and predecessors words $\bm{w}$.
Finally, its output is fed into a fully-connected network and a softmax layer to generate the probability matrix of the output word.

In summary, the whole decoding process is formulated as follows,
\begin{equation}\label{equ:dec1}
  \bm{h}_i = TransD(\bm{w}_{1:i-1}, \bm{m}_{1:T_1}),
\vspace{-0.5cm}
\end{equation}
\begin{equation}\label{equ:dec2}
  \bm{o}_i = Softmax(\mathbf{W}\bm{h}_i + b).
\end{equation}
The conditional probability $p(\bm{s}|\mathbf{V})$ is calculated as follows,
\begin{equation}\label{equ:cond_prob}
  p(\bm{s}|\mathbf{V}) = \prod_{i=1}^{N} p(\bm{s}_i|\bm{s}_{1:i-1}, \mathbf{V}) = \prod_{i=1}^{N} \bm{o}_{i, \bm{s}_i}.
\end{equation}
Finally, the objective function is formulated as follows,
\begin{equation}\label{equ:slt_obj}
  \mathcal{L}_{\mathrm{SLT}} = -\ln p(\bm{s}|\mathbf{V}),
\end{equation}
which is equivalent to calculating the cross-entropy loss on each word.
We adopt the S2T setting~\cite{cihan2018neural}, which directly maps the sign embedding to spoken translation in an end-to-end manner.
During inference, the framework predicts the word one-by-one based on the beam search~\cite{wu2016google}.

For fusion, we leverage the latent visual features from both RGB and pose modalities, \emph{i.e.,} $\mathbf{M}_r$ and $\mathbf{M}_p$, to the same decoder.
Specifically, we replace the original cross attention in the decoder with the cascaded one, which is formulated as follows,
\begin{equation}\label{equ:cas-cross-attn}
    \widetilde{\mathbf{D}}_i = MHA(MHA(\mathbf{Q}_i, \mathbf{M}_p^k, \mathbf{M}_p^v), \mathbf{M}_r^k, \mathbf{M}_r^v),
\end{equation}
where $MHA(\cdot)$ denotes the multi-head cross-attention layer,
$\mathbf{Q}_i$ denotes the feature from the previous decoder layer, and $\widetilde{\mathbf{D}}_i$ is the output of the cascaded attention.

\section{Experiment}
In this section, we first introduce the experiment setup, \emph{i.e.,} datasets, evaluation metrics and implementation details.
Then we perform ablation studies on the framework effectiveness from multiple perspectives.
Finally, we perform extensive experiments to make comparison with state-of-the-art methods on multiple downstream tasks.

\subsection{Experiment Setup}
\subsubsection{Datasets}

We first perform experiments on the dataset with hand pose annotations available to evaluate the framework feasibility.
{HANDS17}~\cite{yuan20172017} is a video-level hand pose estimation dataset with 292,820 frames from 99 video sequences. 
For each video, the first 70\% and remaining 30\% frames are separated for training and testing, respectively.

We evaluate our proposed method on three main video-based sign language understanding tasks, \emph{i.e.,} isolated SLR, continuous SLR and SLT. The corresponding datasets for each task are discussed in the follow.

For \emph{isolated SLR}, we make evlauation on three datasets, \emph{i.e.,} MSASL~\cite{joze2018ms}, WLASL~\cite{li2020word}, and SLR500~\cite{huang2018attention}. MSASL~\cite{joze2018ms} is an American sign language (ASL) dataset containing a vocabulary of 1,000, with 25,512 samples.
{Besides, it also releases two subsets~(MSASL100 and MSASL200) by choosing the Top-$K$ most frequent signs.}
{WLASL~\cite{li2020word} is another ASL dataset with 2,000 signs and 21,083 samples.}
It also contains two subsets, \emph{i.e.,} WLASL100 and WLASL300.
MSASL and WLASL are both collected from the Web, which are recorded in unconstrained real-life conditions with unbalanced samples for each sign word.
These factors bring new challenges on accurate recognition.
{SLR500~\cite{huang2018attention} is the largest CSL dataset, which contains 500 daily signs and 125,000 samples recording at the resolution of $1280\times720$.}
These samples are split into 90,000 and 35,000 for training and testing, respectively.

For \emph{continuous SLR}, the evaluation is conducted on two datasets, \emph{i.e.,} RWTH-Phoenix~\cite{koller2015continuous} and RWTH-PhoenixT~\cite{cihan2018neural}.  RWTH-Phoenix~\cite{koller2015continuous} is a popular German sign language dataset collected from the weather forecast broadcast. 
It contains 6,841 samples, with 5,672, 540 and 629 videos for training, validation and testing, respectively.
RWTH-PhoenixT~\cite{cihan2018neural} is included for evaluation, which is introduced in ``SLT datasets''.

For \emph{SLT}, we make evaluation on RWTH-PhoenixT~\cite{cihan2018neural} dataset, which is treated as the extended version of RWTH-Phoenix.
It provides parallel sign gloss and translation annotations, to evaluate both continuous SLR and SLT tasks. 
It contains 8,257 videos, which are divided into three sets: 7,096 for training, 519 for validation, and 642 for testing.
RWTH-Phoenix and RWTH-PhoenixT are both recorded at the resolution of $210\times260$.

\subsubsection{Evaluation Metrics}
To evaluate whether our framework works during pre-training, we adopt the metrics for evaluating pose estimation accuracy.
Specifically, we report the Percentage of Correct Keypoints~(PCK) score and the Area Under the Curve~(AUC) on the PCK threshold ranging from 20 to 40 pixels.
PCK defines the keypoint to be correct if the Euclidean distance between this keypoint and ground truth is lower than the threshold.
The distance metric is expressed in pixels.

For \emph{isolated SLR}, We utilize the accuracy metrics, including per-instance~(P-I) and per-class~(P-C) metrics.
P-I and P-C denote the average accuracy over all the instances and classes, respectively.
Following previous works~\cite{albanie2020bsl,hu2021signbert}, we report Top-1 and Top-5 P-I and P-C metrics on MSASL and WLASL.
Since each class in SLR500 contains the same number of samples, P-I is equal to P-C and we only report one of them.

For \emph{continuous SLR}, we utilize Word Error Rate~(WER) as the evaluation metric.
WER is the editing distance, which measures the least operations~(substitution, deletion and insertion) to transform the hypothesis to the reference gloss sentence as follows,
\begin{equation}\label{equ:wer}
  WER = \frac{N_i + N_d + N_s}{L},
\end{equation}
where $N_i$, $N_d$, and $N_s$ are the number of operations for insertion, deletion, and substitution, respectively.
$L$ denotes the length of the reference sentence.

For \emph{SLT}, we adopt BLEU~\cite{papineni2002bleu} and ROUGE~\cite{lin2004rouge} metrics which are commonly utilized in neural machine translation.
BLEU calculates the overlap rate of $n$-gram between the generated text and the reference text, and $n$ ranges from 1 to 4.
ROUGE is a metric based on the recall rate and measures the sentence-level structure similarity.
In this work, we refer to the ROUGE-L F1-Score.

\subsubsection{Implementation Details}
In our experiment, all the models are implemented by PyTorch~\cite{paszke2019pytorch} and trained on NVIDIA RTX 3090.
Since sign language datasets contain no available pose annotations, we utilize MMPose~\cite{mmpose2020} to extract 133 full 2D keypoints, consisting of 23 body joints, 68 face joints and 42 hand joints.
The hyper-parameters $\epsilon$, $\lambda$, $w_{\beta}$ and $w_{\delta}$ are set as 0.5, 0.01, 10.0 and 100.0, respectively.
During decoding, the beam width is set as 10 and 3 for continuous SLR and SLT, respectively.

During the pre-training stage, the utilized data includes the training data from all aforementioned sign datasets, along with other collected data from~\cite{hu2020global,duarte2021how2sign}.
In total, the pre-training data volume is 230,246 videos.
The Adam optimizer is adopted to train the framework for 60 epochs with the weight decay set as 0.01.
The learning rate warms up over the first 10\% of the training process, and then decays linearly from the peak rate~(1e-4).
All the frames are fed into the framework.

For all the downstream tasks, the Adam optimizer is still adopted.
The learning rate for isolated SLR, continuous SLR and SLT are 1e-4, 1e-4 and 5e-5, respectively.
For continuous SLR and SLT, we follow the setting\cite{camgoz2020sign}.
Spatial-temporal data augmentation is utilized during training.
\emph{Spatially}, following~\cite{yan2018spatial}, we adopt random moving augmentation to simulate spatial disturbance induced by rotation, translation and scaling factors.
\emph{Temporally}, for isolated SLR, we extract 32 frames from the origin video using random and center sampling for training and testing, respectively.
While for continuous SLR and SLT, we randomly
sample 80\% frames during training and utilize all the frames during testing.

\begin{table}[!t]
  \small
  \tabcolsep=7.5pt
  \caption{Framework feasibility validation on HANDS17. 
  ``P@20'' denotes the PCK metrics with the error threshold set as 20 pixel. 
  ``Joint'', ``Frame'' and ``Clip'' denote the masked joint modeling, masked frame modeling and masked clip modeling, respectively.
  ``Input'' and ``Output'' represent the corrupted input pose and the reconstructed pose sequence by our framework, respectively.}
  \begin{center}
   \begin{tabular}{ccc|cc|cc}
    \hline
    \multicolumn{3}{c|}{Mask} & \multicolumn{2}{c|}{Input} & \multicolumn{2}{c}{Output} \\
    Joint       & Frame       &  Clip      & P@20  & AUC   & P@20  & AUC   \\ \hline \hline
    \checkmark  &             &            & 90.06 & 89.99 & 94.60 & 95.15 \\
                & \checkmark  &            & 74.83 & 74.80 & 93.30 & 95.13 \\
                &             & \checkmark & 60.99 & 60.00 & 91.94 & 93.47 \\
    \checkmark  & \checkmark  & \checkmark & 66.65 & 66.63 & 94.00 & 94.74 \\ \hline          
   \end{tabular}
  \end{center}
  \label{tab:HANDS17}
  \vspace{-0.2cm}
\end{table}

\begin{table}[!t]
  \small
  \tabcolsep=8.5pt
  \caption{Impact of the Transformer layers $N$ on MSASL dataset. 
  $N$ denotes the number of the Transformer encoder layers in our framework.}
  \begin{center}
  \begin{tabular}{c|cc|cc|cc}
  \hline
  \multicolumn{1}{c|}{\multirow{2}{*}{$N$}} & \multicolumn{2}{c|}{100} & \multicolumn{2}{c|}{200} & \multicolumn{2}{c}{1000} \\
        & P-I & P-C & P-I & P-C & P-I & P-C   \\ \hline \hline
  2 & 82.56 & 82.35  & 74.47 & 75.51 & 59.20 & 56.70   \\
  3 & \textbf{84.94} & \textbf{85.23} & \textbf{78.51} & \textbf{79.35} & \textbf{62.42} & \textbf{60.15} \\ 
  4 & 83.75 & 83.56 & 76.97 & 77.74 & 60.69 & 57.34  \\
  5 & 83.88 & 84.23 & 77.04 & 77.93 & 61.27 & 58.30  \\ \hline
  \end{tabular}
  \end{center}
  \label{tab:num-layer}
  \vspace{-0.2cm}
  \end{table}

\subsection{Ablation Study}
In this section, we first validate the framework feasibility.
Then we perform detailed ablation studies on different components of our framework.

\subsubsection{Framework Feasibility}
\begin{figure*}
    \centering
    \includegraphics[width=1.0\linewidth]{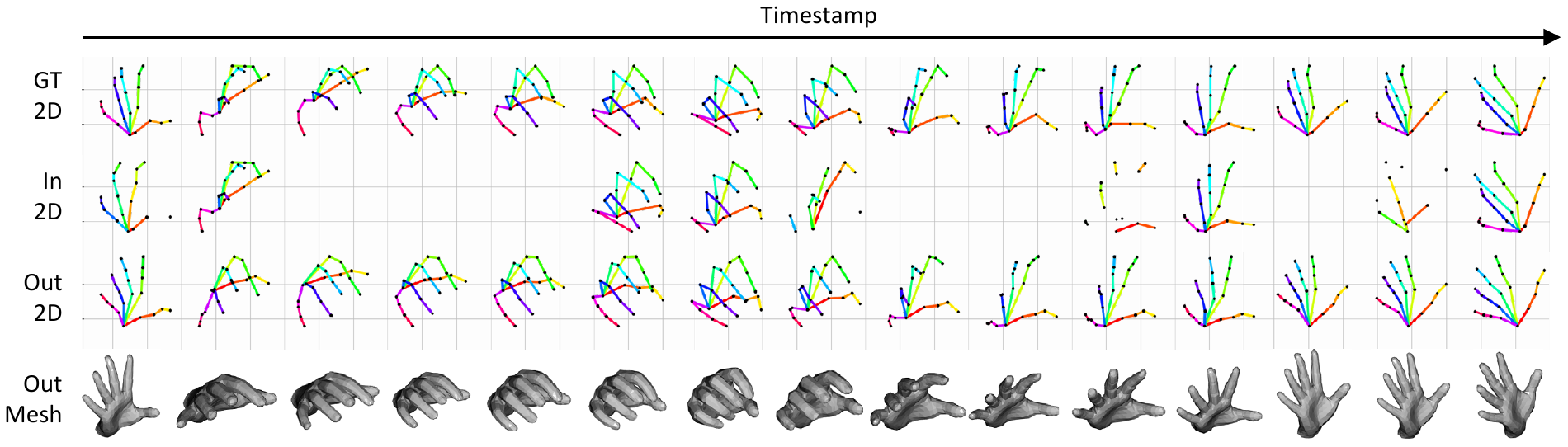}
    \vspace{-0.3cm}
    \caption{Qualitative illustration of the framework feasibility on HANDS17.
    We exhibit 15 continuous frames of a video.
    Four rows represent the ground truth pose, input pose disturbed by all kinds of masked modeling strategies~(joint, frame and clip), reconstructed sequence and the middle mesh representation, respectively.
    Notably, blanks in the second row represent all joints in the corresponding frames are masked.
    }
    \label{fig:HANDS17}
  \end{figure*}

We validate the framework feasibility via observing its pose reconstruction capability, on HANDS17 dataset with hand pose annotation available.
In this setting, we adopt all masked modeling strategies to train our framework on this dataset.
During validation, we perform different masking cases on the input sequence and evaluate the framework output quality.
As shown in Table~\ref{tab:HANDS17}, the output metrics are higher than those of the input under all masking cases, which validates our framework feasibility.
Besides, we qualitatively visualize the hand pose reconstruction in Figure~\ref{fig:HANDS17}. 
It can be observed that the reconstructed hand pose sequence is consistent with the ground truth, even under the severely corrupted input situation.
It is largely attributed to inherent contextual cues captured by our framework via our designed pretext task.

\subsubsection{Ablation Study}
To study the impact of different hyper-parameters and settings in our approach, we conduct experiments on MSASL and its subset with per-instance and per-class Top-1 accuracy as the performance indicator.

\noindent \textbf{Impact of the Transformer layers $N$.}
As shown in Table~\ref{tab:num-layer}, the accuracy gets improved when the number $N$ of Transformer layers increases.
It reaches the peak when $N$ is equal to 3.
There exists difference in the best $N$ between the original BERT and ours, which may be attributed to different characteristics between the sign pose and text domains.
In all the experiments, we set $N$ as 3 unless stated.

\noindent \textbf{Impact of the pose $\theta$ dimension in the hand-model-aware decoder.}
From Table~\ref{tab:pose-num}, the pose $\theta$ dimension represents the MANO characterization capability of the hand gesture.
The increase of the pose dimension brings enhanced capability and accuracy improvement on the downstream SLR. 
It reaches the top when the dimension is equal to 25.
However, further increasing does not bring more performance gain, which may be caused by the optimization difficulty.

\noindent \textbf{Impact of the temporal span $K$ in masked clip modeling.}
In Table~\ref{tab:temporal_span}, it can be observed that the accuracy reaches the top when $K$ is equal to 8.
The suitable temporal mask span enforces the framework to capture the temporal dynamics during sign language.
In the following, we set $K$ as 8 unless stated.

\begin{table}[!t]
  \small
  \tabcolsep=5pt
  \caption{Impact of the pose $\bm{\theta}$ dimension in the hand-model-aware decoder on MSASL dataset.}
  \label{tab:pose-num}
  \begin{center}
  \begin{tabular}{c|cc|cc|cc}
  \hline
  \multicolumn{1}{c|}{\multirow{2}{*}{Dimension}} & \multicolumn{2}{c|}{100} & \multicolumn{2}{c|}{200} & \multicolumn{2}{c}{1000} \\
            & P-I & P-C & P-I   & P-C   & P-I   & P-C   \\ \hline \hline
  15 & 82.83 & 82.83 & 76.01 & 76.50 & 61.65 & 58.59   \\
  25 & \textbf{84.94} & \textbf{85.23} & \textbf{78.51} & \textbf{79.35} & \textbf{62.42} & \textbf{60.15} \\  
  35 & 83.88 & 84.20 & 77.19 & 77.99 & 61.60 & 59.04   \\ \hline 
  \end{tabular}
  \end{center}
  \vspace{-0.2cm}
\end{table}

\begin{table}[!t]
  \small
  \tabcolsep=8pt
  \caption{Impact of the temporal span $K$ in masked clip modeling on MSASL dataset. $K$ represents that the masked clip duration ranges from 2 to $K$.}
  \begin{center}
  \begin{tabular}{c|cc|cc|cc}
  \hline
  \multicolumn{1}{c|}{\multirow{2}{*}{$K$}} & \multicolumn{2}{c|}{100} & \multicolumn{2}{c|}{200} & \multicolumn{2}{c}{1000} \\
        & P-I & P-C & P-I & P-C & P-I & P-C   \\ \hline \hline
  4 & 81.90 & 82.17 & 73.58 & 74.17 & 60.43 & 57.51   \\
  8 & \textbf{83.88} & \textbf{83.55} & \textbf{76.60} & \textbf{77.57} & \textbf{61.94} & \textbf{59.76}  \\  
  12 & 82.96 & 82.83 & 74.98 & 75.16 & 60.93 & 58.67  \\ \hline
  \end{tabular}
  \end{center}
  \label{tab:temporal_span}
  \vspace{-0.3cm}
\end{table}

\begin{table}[!t]
  \small
  \tabcolsep=5pt
  \caption{{Impact of different temporal information extraction on MSASL dataset.}}
  \label{tab:tem_info}
  \begin{center}
  \begin{tabular}{l|cc|cc|cc}
  \hline
  \multicolumn{1}{c|}{\multirow{2}{*}{Method}} & \multicolumn{2}{c|}{100} & \multicolumn{2}{c|}{200} & \multicolumn{2}{c}{1000} \\
            & P-I & P-C & P-I   & P-C   & P-I   & P-C   \\ \hline \hline
  PE & \textbf{84.94} & \textbf{85.23} & \textbf{78.51} & \textbf{79.35} & \textbf{62.42} & \textbf{60.15} \\ 
  GCN\_Tem-3 & 83.36 & 83.41 & 76.09 & 76.85 & 60.21 & 57.62  \\
  GCN\_Tem-5 & 83.62 & 84.36 & 76.09 & 77.12 & 60.55 & 58.13 \\ \hline
  \end{tabular}
  \end{center}
  \vspace{-0.3cm}
\end{table}

\noindent {\textbf{Impact of different temporal information extraction on MSASL dataset.}
There are many alternative methods to extract temporal information.
Besides temporal position encoding, directly extracting temporal information is also a solution.
To this end, we modify the current GCN into the temporal variant following the practice in~\cite{cai2019exploiting}.
Specifically, the original spatial GCN graph is replaced with the spatial-temporal one via adding the local connections along the temporal dimension.
With this modification, the gesture state extractor embeds the temporal receptive field of additional $k$ adjacent input frames and thus captures the temporal information.
Besides, since our pretext task needs to recover the masked pose token in the corresponding output, we utilize padding to keep the sequence length after the gesture extractor the same as the input.}

{As shown in Table~\ref{tab:tem_info}, we perform comparison on these different temporal extraction methods.
``PE'' denotes utilizing the position encoding for temporal information extraction.
For ``GCN\_Tem-$k$'', we remove the temporal position encoding and directly extract temporal information via our modified GCN backbone.
$k$ represents the number of adjacent frames.
These settings achieve comparable performance on the downstream SLR.
It can be explained that the following Transformer encoder contains the strong capability of capturing long-term sequential dependencies.
The simple position encoding is sufficient to indicate this encoder with the temporal order.
Unless stated, we utilize the temporal position encoding as the temporal order indicator.}

\begin{table}[!t]
  \small
  \tabcolsep=4pt
  \caption{Effectiveness of the spatial-temporal position encoding~(``PE'') on MSASL dataset.}
  \begin{center}
  \begin{tabular}{cc|cc|cc|cc}
  \hline
  \multicolumn{2}{c|}{PE} & \multicolumn{2}{c|}{100} & \multicolumn{2}{c|}{200} & \multicolumn{2}{c}{1000} \\
  Temporal   & Spatial       & P-I   & P-C    & P-I   & P-C   & P-I   & P-C \\ \hline \hline
              &  & 77.68 & 77.93 & 73.07 & 73.50 & 51.80 & 48.76 \\
  \checkmark &            & 79.00 & 79.05 & 74.39 & 74.51 & 52.97 & 49.95 \\
              &  \checkmark & 82.96 & 82.75 & 75.94 & 77.15 & 60.95 & 58.82  \\
  \checkmark & \checkmark & \textbf{84.94} & \textbf{85.23} & \textbf{78.51} & \textbf{79.35} & \textbf{62.42} & \textbf{60.15} \\ 
  \hline
  \end{tabular}
  \vspace{-1em}
  \end{center}
  \label{tab:st-pe}
  \end{table}

\begin{table}[!t]
  \small
  \tabcolsep=8pt
  \caption{Effectiveness of the masking ratio $R$ on MSASL dataset.}
  \begin{center}
  \begin{tabular}{l|cc|cc|cc}
  \hline
  \multicolumn{1}{l|}{\multirow{2}{*}{R}} & \multicolumn{2}{c|}{100} & \multicolumn{2}{c|}{200} & \multicolumn{2}{c}{1000} \\
              & P-I & P-C & P-I   & P-C   & P-I   & P-C   \\ \hline \hline
  20\%  & 78.47 & 78.46 & 69.09 & 70.62 & 59.52 & 56.15 \\
  30\% & 82.30 & 82.33 & 76.23 & 76.79 & 61.07 & 57.92  \\
  40\% & \textbf{84.94} & \textbf{85.23} & \textbf{78.51} & \textbf{79.35} & \textbf{62.42} & \textbf{60.15}  \\
  50\% & 82.96 & 83.26 & 76.31 & 77.40 & 60.93 & 57.95  \\
  60\% & 80.18 & 81.06 & 75.64 & 76.21 & 59.80 & 57.25 \\ \hline   
  \end{tabular}
  \end{center}
  \label{tab:mask-ratio}
  \end{table}

  \begin{table}[!t]
  \small
  \tabcolsep=3.8pt
  \caption{Effectiveness of the masking strategy on MSASL dataset. The first row denotes the baseline, \emph{i.e.,} our framework is trained without pre-training. ``Joint'', ``Frame'' and ``Clip'' denote the masked joint modeling, masked frame modeling and masked clip modeling, respectively.}
  \begin{center}
  \begin{tabular}{ccc|cc|cc|cc}
  \hline
  \multicolumn{3}{c|}{Mask} & \multicolumn{2}{c|}{100} & \multicolumn{2}{c|}{200} & \multicolumn{2}{c}{1000} \\
  Joint      & Frame      & Clip       & P-I   & P-C    & P-I   & P-C   & P-I   & P-C \\ \hline \hline
              &            &            & 78.20 & 78.10 & 72.77 & 73.21 & 53.31 & 50.43  \\
  \checkmark &            &            & 82.03 & 82.78 & 74.39 & 74.62 & 58.44 & 55.24 \\
              & \checkmark &            & 82.69 & 82.85 & 76.53 & 77.24 & 60.23 & 58.02 \\
              &            & \checkmark & 83.88 & 83.55 & 76.60 & 77.57 & 61.94 & 59.76 \\
  \checkmark & \checkmark & \checkmark & \textbf{84.94} & \textbf{85.23} & \textbf{78.51} & \textbf{79.35} & \textbf{62.42} & \textbf{60.15} \\ \hline          
  \end{tabular}
  \end{center}
  \label{tab:mask}
  \end{table}

\noindent \textbf{Effectiveness of the spatial-temporal position encoding.}
As shown in Table~\ref{tab:st-pe}, we validate the effectiveness of spatial-temporal position encoding.
The first row denotes the results without any position encoding.
In the second row, we incorporate the temporal inforamtion with the temporal position encoding.
{In the third row, we incorporate the global hand information in the form of hand spatial position encoding and involve all arm joints~(totally 7 joints).
Compared with the temporal counterpart, our designed spatial position encoding incorporates the important hand global position information and brings a larger performance gain.}
Furthermore, these two encoding schemes exhibit complementary effects, bringing even 10.62\% per-instance Top-1 accuracy improvement over the baseline on the full set.

\noindent \textbf{Effectiveness of the masking ratio $R$.}
We study the influence of the masking ratio in Table~\ref{tab:mask-ratio}.
It is observed that the performance reaches the top when the masking ratio is equal to 40\%, which is inconsistent with BERT~\cite{devlin2018bert}.
It may be attributed to the information density difference between sign pose and predefined NLP word tokens.

\begin{table}[!t]
  \small
  \tabcolsep=6pt
  \caption{Effectiveness of the model-aware decoder on MSASL dataset. We compare ours with different pose decoders.}
  \begin{center}
  \begin{tabular}{l|cc|cc|cc}
  \hline
  \multicolumn{1}{l|}{\multirow{2}{*}{Decoder}} & \multicolumn{2}{c|}{100} & \multicolumn{2}{c|}{200} & \multicolumn{2}{c}{1000} \\
              & P-I & P-C & P-I   & P-C   & P-I   & P-C   \\ \hline \hline
  1-layer fc & 82.56 & 83.11 & 74.32 & 75.26 & 59.88 & 57.17 \\
  2-layer fc & 83.62 & 83.82 & 75.06 & 75.92 & 60.50 & 57.65   \\
  Ours       & \textbf{84.94} & \textbf{85.23} & \textbf{78.51} & \textbf{79.35} & \textbf{62.42} & \textbf{60.15} \\ \hline   
  \end{tabular}
  \end{center}
  \label{tab:model-aware}
  \end{table}

  \begin{table}[!t]
  \small
  \tabcolsep=7.5pt
  \caption{Effectiveness of the ratio of pre-training data scale on MSASL dataset.}
  \begin{center}
  \begin{tabular}{l|cc|cc|cc}
  \hline
  \multicolumn{1}{l|}{\multirow{2}{*}{Ratio}} & \multicolumn{2}{c|}{100} & \multicolumn{2}{c|}{200} & \multicolumn{2}{c}{1000} \\
              & P-I & P-C & P-I   & P-C   & P-I   & P-C   \\ \hline \hline
  0\%  & 78.20 & 78.10 & 72.77 & 73.21 & 53.31 & 50.43 \\
  25\% & 81.11 & 81.44 & 74.83 & 75.51 & 58.53 & 56.04  \\
  50\% & 81.24 & 81.79 & 75.35 & 76.35 & 59.61 & 56.71  \\
  75\% & 83.09 & 83.36 & 76.01 & 76.60 & 60.43 & 57.63  \\
  100\% & \textbf{84.94} & \textbf{85.23} & \textbf{78.51} & \textbf{79.35} & \textbf{62.42} & \textbf{60.15} \\ \hline   
  \end{tabular}
  \end{center}
  \vspace{-1em}
  \label{tab:data-scale}
  \end{table}

  \begin{figure*}[t!]
    \centering
    \includegraphics[width=1.0\linewidth]{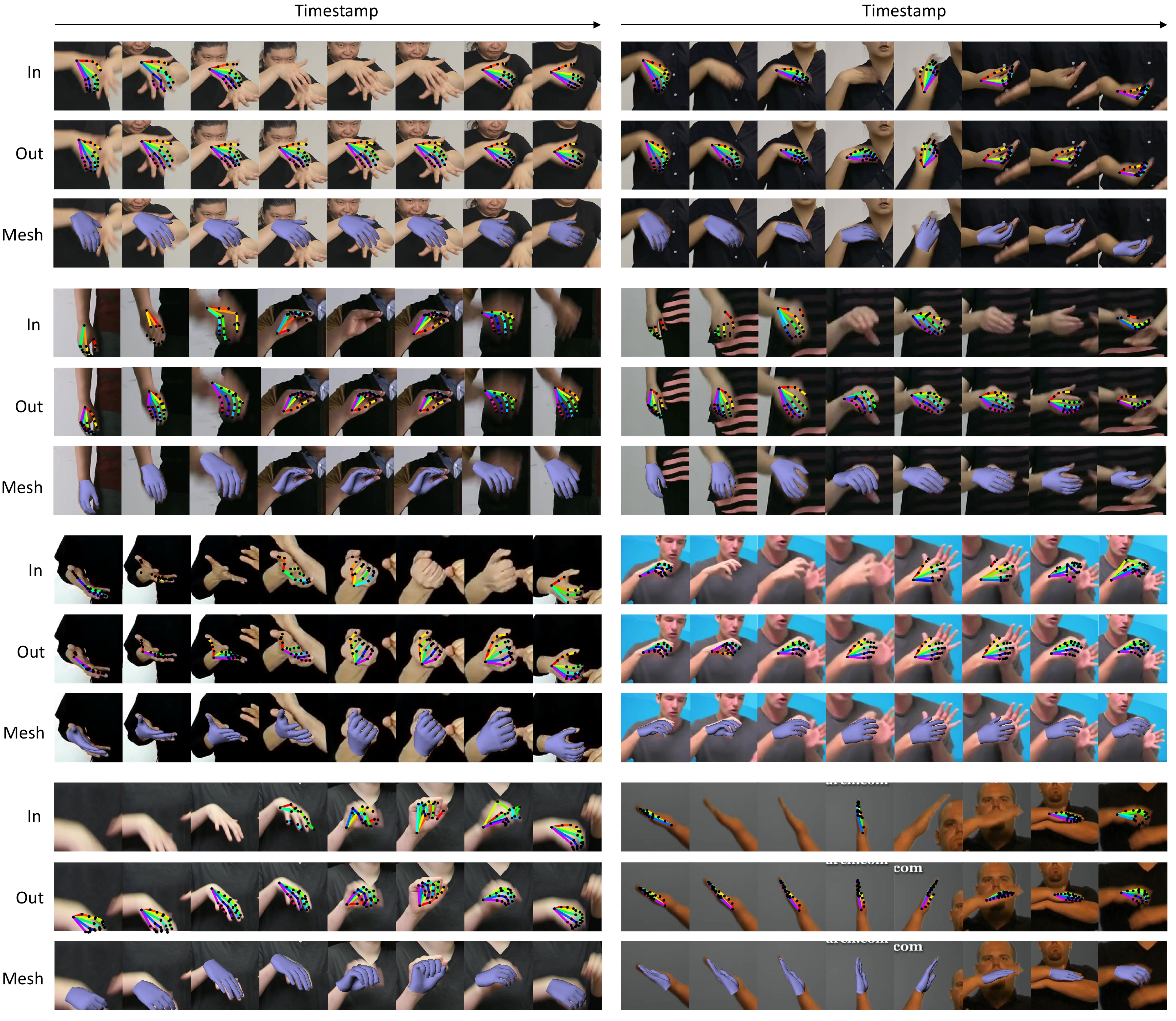}
    \caption{More visualization samples under sign data sources with no hand pose annotation. 
    For each sample, 8 continuous frames are visualized. 
    ``In'', ``Out'' and ``Mesh'' denote the input hand pose, the reconstructed hand pose and the intermediate hand mesh, respectively. 
    For clarity, we visualize all the poses and meshes on their aligned RGB image planes.}
    \label{fig:vis_more}
  \end{figure*}
  
  \begin{figure*}[t!]
    \centering
    \includegraphics[width=1.0\linewidth]{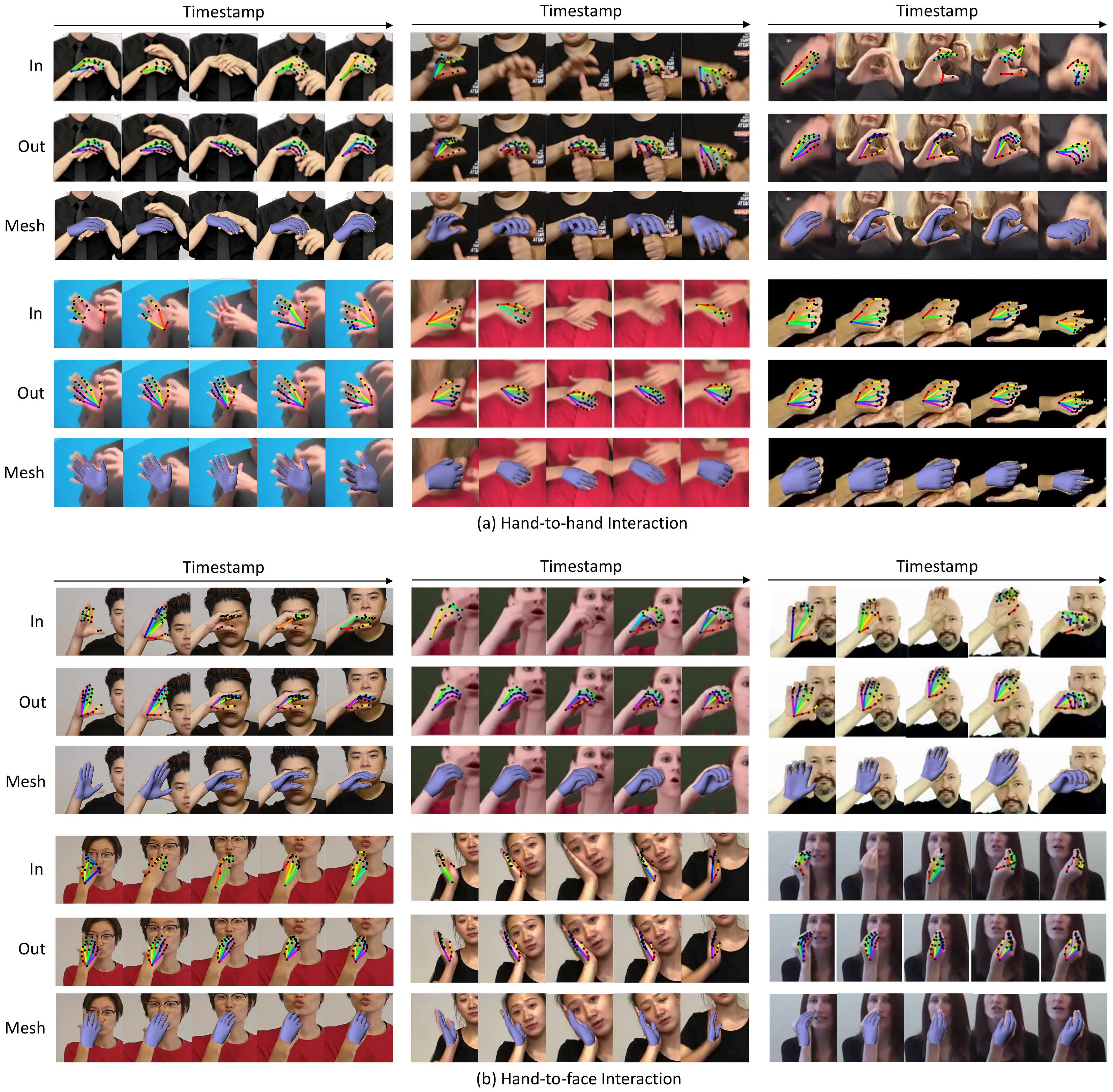}
    \caption{{More visualization samples on two types of hard interaction cases during sign language expression, \emph{i.e.,} hand-to-hand interaction and hand-to-face interaction. 	
    For each sample, 5 continuous frames are visualized. 
    ``In'', ``Out'' and ``Mesh'' denote the input hand pose, the reconstructed hand pose and the intermediate hand mesh, respectively. 
    For clarity, we only plot one hand and visualize its poses and meshes on their aligned RGB image planes.}}
    \vspace{-0.2cm}
    \label{fig:vis_more2}
  \end{figure*}

\noindent \textbf{Effectiveness of the masking strategy.}
As demonstrated in Table~\ref{tab:mask}, the first row denotes the baseline method, which is fine-tuned on MSASL without pre-training.
Our designed three masked modeling strategies target at different levels of context contained in the sign language domain, which individually bring 5.13\%, 6.92\% and 8.63\% per-instance Top-1 accuracy improvement over the baseline on the full set, respectively.
Among them, masked clip modeling brings the largest performance gain.
When all these masking strategies are utilized, it reaches the best performance.

\noindent \textbf{Effectiveness of the model-aware decoder.}
We compare the effects of different pose decoders in Table~\ref{tab:model-aware}.
The first two rows represent that we utilize fully-connected layers to directly regress the keypoints for pose reconstruction during pre-training.
Compared with the direct regression method, our decoder incorporates prior via regressing the compact gesture embedding, which eases optimization and brings a larger performance gain on the downstream SLR.
Besides, our decoder also exhibits the additional benefit, which inflates the input 2D sequence to the corresponding 3D mesh as the intermediate representation.

\noindent \textbf{Effectiveness of the pre-training data scale.}
We study the impact of the pre-training data scale in Table~\ref{tab:data-scale}.
We randomly extract a portion of all training data, indicated in the ``Ratio'' column.
Then we pre-train and fine-tune our framework based on the same setup.
It is observed that the accuracy grows monotonically when the pre-training data volume increases, which suggests our framework may benefit from even more pre-training data.

\begin{table*}[!t]
  \small
  \tabcolsep=4pt
  \caption{Comparison with other pre-training strategies on downstream tasks.
  For fair comparison, all the pre-training methods are performed on the same backbone, \emph{i.e.,} embedding layer and Transformer encoder.
  The first row represents the framework is directly fine-tuned on the downstream tasks without pre-training.
  ``Partial'' and ``All'' denote the corresponding classification data and all pre-training data, respectively.
  The data volumes of ``Partial'' and ``All'' are about 160k and 230k videos, respectively.
  ($\uparrow$ denotes the higher the better, while $\downarrow$ represents the lower the better.)
  }
  \begin{center}
  \begin{threeparttable}
  \resizebox{\linewidth}{!}{
  \begin{tabular}{l|l|cc|cc|cc|cc|c|cc|cc}
  \hline
  \multirow{3}{*}{Method} & \multirow{3}{*}{Pre-Train} & \multicolumn{4}{c|}{MSASL}
                          & \multicolumn{4}{c|}{WLASL}
                          & \multicolumn{1}{c|}{SLR500} & \multicolumn{2}{c|}{RWTH-Phoenix} & \multicolumn{2}{c}{RWTH-PhoenixT}  \\ \cline{3-15}
          & & \multicolumn{2}{c|}{P-I} & \multicolumn{2}{c|}{P-C} 
          & \multicolumn{2}{c|}{P-I} & \multicolumn{2}{c|}{P-C}
          & \multicolumn{1}{c|}{P-I} & Dev & Test & Dev & Test \\ 
          & & Top-1 $\uparrow$ & Top-5 $\uparrow$ & Top-1 $\uparrow$ & Top-5 $\uparrow$ 
          & Top-1 $\uparrow$ & Top-5 $\uparrow$ & Top-1 $\uparrow$ & Top-5 $\uparrow$
          & Top-1 $\uparrow$ & WER $\downarrow$ & WER $\downarrow$ & WER $\downarrow$ & WER $\downarrow$    \\ \hline \hline
  Baseline & Scratch & 53.31 & 75.98 & 50.43 & 74.12 & 38.33 & 72.59 & 36.40 & 71.23 & 92.1 & 43.6 & 43.4 & 42.2 & 42.6 \\ \hline
  Supervised & Partial & 58.82 & 80.42 & 56.27 & 79.54 & 46.00 & 79.95 & 43.63 & 78.32 & 94.7 & 42.5 & 42.6 & 40.8 & 41.7 \\
  V-MoCo~\cite{feichtenhofer2021large} & Partial & 54.22 & 78.26 & 51.31 & 77.03 & 39.12 & 72.79 & 36.93 & 71.15 & 94.1 & 42.1 & 43.4 & 40.3 & 40.8 \\
  Ours & Partial & \textbf{60.13} & \textbf{82.12} & \textbf{57.19} & \textbf{80.72} & \textbf{47.46} & \textbf{81.62} & \textbf{45.03} & \textbf{80.31} & \textbf{95.1} & \textbf{36.7} & \textbf{36.2} & \textbf{35.0} & \textbf{35.2} \\ \hline 
  V-MoCo~\cite{feichtenhofer2021large} & All & 55.27 & 79.72 & 52.06 & 78.31 & 40.17 & 75.19 & 37.71 & 73.38 & 94.8 & 41.1 & 41.3 & 39.2 & 40.0 \\
  Ours & All & \textbf{62.42} & \textbf{83.49} & \textbf{60.15} & \textbf{82.44} & \textbf{48.85} & \textbf{82.48} & \textbf{46.37} & \textbf{81.33} & \textbf{95.4} & \textbf{34.0} & \textbf{34.1} & \textbf{32.9} & \textbf{33.6} \\ \hline   \noalign{\smallskip} \noalign{\smallskip}
  \end{tabular}
  }
  \resizebox{\linewidth}{!}{
  \begin{tabular}{l|l|ccccc|ccccc}
    \hline
    \multirow{3}{*}{Method} & \multirow{3}{*}{Pre-Train} & \multicolumn{10}{c}{RWTH-PhoenixT}\\ \cline{3-12}
            & & \multicolumn{5}{c|}{Dev} & \multicolumn{5}{c}{Test} \\ 
            & & ROUGE $\uparrow$ & BLEU-1 $\uparrow$ & BLEU-2 $\uparrow$ & BLEU-3 $\uparrow$ & BLEU-4 $\uparrow$
            & ROUGE $\uparrow$ & BLEU-1 $\uparrow$ & BLEU-2 $\uparrow$ & BLEU-3 $\uparrow$ & BLEU-4 $\uparrow$ \\ \hline \hline
    Baseline   & Scratch & 40.23 & 39.09 & 26.40 & 19.63 & 15.50 & 39.83 & 39.27 & 26.98 & 20.10 & 15.90 \\ \hline
    Supervised & Partial & 41.63 & 40.87 & 28.06 & 20.84 & 16.49 & 41.58 & 41.60 & 28.82 & 21.68 & 17.29 \\
    V-MoCo~\cite{feichtenhofer2021large} & Partial & 42.44 & 41.94 & 28.97 & 21.43 & 16.84 & 41.75 & 41.83 & 28.82 & 21.36 & 16.82 \\
    Ours & Partial & \textbf{44.15} & \textbf{44.03} & \textbf{31.18} & \textbf{23.71} & \textbf{19.00} & \textbf{44.33} & \textbf{43.51} & \textbf{30.92} & \textbf{23.59} & \textbf{19.10} \\ \hline 
    V-MoCo~\cite{feichtenhofer2021large} & All & 42.79 & 42.55 & 29.40 & 22.13 & 17.53 & 41.95 & 42.68 & 29.49 & 21.89 & 17.36 \\
    Ours & All & \textbf{45.53} & \textbf{44.45} & \textbf{31.88} & \textbf{24.59} & \textbf{19.86} & \textbf{44.89} & \textbf{44.35} & \textbf{32.09} & \textbf{24.92} & \textbf{20.41} \\ \hline   
    \end{tabular}
  }
  \end{threeparttable}
  \end{center}
  \vspace{-1em}
  \label{tab:ptm}
\end{table*}

\subsection{Qualitative Visualization}
{We visualize the reconstruction results under the sign data sources after pre-training in Figure~\ref{fig:vis_more} and \ref{fig:vis_more2}.
In this setting, poses are detected by an off-the-shelf extractor and fed into the framework.
For clarity, we only plot one hand and visualize its reconstructed poses and the intermediate hand meshes, which are produced by our framework.
We start by demonstrating some general cases in Figure~\ref{fig:vis_more}.
Under loss of hand joints in several frames, our framework is able to capture the context to reconstruct the pose and mesh, which align the 2D image plane well.
The hand mesh, as the intermediate representation, also improves the interpretability of our method.}

{As illustrated in Figure~\ref{fig:vis_more2}, we further demonstrate qualitative results on two types of hard cases, \emph{i.e.,} hand-to-hand interaction and hand-to-face interaction.
Due to the similar appearance of hand and face and complex self- and mutual occlusion, these interactions bring inherent ambiguity and cause failure in hand pose estimation.
Even under these hard cases, our framework can rectify the noisy inputs and infer all the poses which well align the image plane.
This strong hallucination capability may be largely attributed to the well-modeled statistics in the sign language domain.}

\subsection{Comparison with Other Pre-Training Strategies}
As demonstrated in Table~\ref{tab:ptm}, we compare with other pre-training strategies, including supervised and state-of-the-art self-supervised pre-training methods.
For fair comparison, we pre-train on the same SignBERT+ encoder backbone~(embedding layer and Transformer encoder).

Similar to~\cite{feichtenhofer2021large}, supervised pre-training denotes that we add a classifier~(an MLP and softmax layer) on top of the backbone and perform pre-training under the classification task.
Specifically, supervised pre-training is conducted on a portion of the original pre-training data~(denoted as ``Partial''), \emph{i.e.,} the corresponding classification~(isolated SLR) benchmarks.
The ``Partial'' and ``All'' data volumes are 160,113 and 230,246 videos, respectively.

For fair comparison with supervised pre-training, we define two evaluation settings for self-supervised methods, \emph{i.e.,} pre-training on the ``Partial'' and ``All'' data volumes.
We adopt the state-of-the-art self-supervised pre-training method, \emph{i.e.,} V-MoCo~\cite{feichtenhofer2021large}.
It is based on contrastive learning, and can be treated as the extended version of MoCo~\cite{he2020momentum} into the video domain.
Since it originally works on the RGB domain, we make a few modifications by replacing its backbone with ours, which is able to process the pose modality.
During pre-training, we randomly sample two clips with 32 consecutive frames from the same sign pose sequence, as the query and positive samples.
The negative samples are obtained from the clips of other videos.
During training, its objective is to maximize the similarity between the query and positive samples, which aims to learn temporally persistent features of the same video.

We evaluate the effectiveness of these pre-training methods on all three downstream tasks, \emph{i.e.,} isolated SLR~(MSASL, WLASL, SLR500), continuous SLR~(RWTH-Phoenix and RWTH-PhoenixT) and SLT~(RWTH-PhoenixT).
In isolated SLR, supervised pre-training brings a larger performance gain than V-MoCo.
While for continuous SLR and SLT, the supervised pre-training method brings a relatively smaller performance gain, whose performance is worse than that of V-MoCo.
Supervised pre-training exhibits the limited generalization capability to the downstream tasks, whose objective is inconsistent with classification.
For self-supervised pre-training methods, \emph{i.e.,} V-MoCo and Ours, they do not rely on annotated data and scale well with larger pre-training data volume.
Notably, when compared with other pre-training strategies, our method achieves the best performance on all downstream tasks with notable gains.

  \begin{table*}[!t]
    \small
    \tabcolsep=6.2pt
    \caption{Evaluation of isolated SLR on MSASL dataset~(the higher the better). \cite{joze2018ms} denotes the RGB baseline for fusion.}
    \begin{center}
    \begin{threeparttable}
    \begin{tabular}{l|cc|cc|cc|cc|cc|cc}
    \hline
    \multirow{3}{*}{Method} & \multicolumn{4}{c|}{MSASL100}
                            & \multicolumn{4}{c|}{MSASL200}
                            & \multicolumn{4}{c}{MSASL}\\ \cline{2-13}
            & \multicolumn{2}{c|}{Per-instance} & \multicolumn{2}{c|}{Per-class} 
            & \multicolumn{2}{c|}{Per-instance} & \multicolumn{2}{c|}{Per-class}
            & \multicolumn{2}{c|}{Per-instance} & \multicolumn{2}{c}{Per-class} \\ 
            & Top-1 & Top-5 & Top-1 & Top-5 
            & Top-1 & Top-5 & Top-1 & Top-5 
            & Top-1 & Top-5 & Top-1 & Top-5    \\ \hline \hline
    \textbf{Pose-based} & & & & & & & & & & & & \\
    ST-GCN~\cite{yan2018spatial}  & 59.84 & 82.03 & 60.79 & 82.96  
            & 52.91 & 76.67 & 54.20 & 77.62
            & 36.03 & 59.92 & 32.32 & 57.15 \\
    SignBERT~\cite{hu2021signbert} & 76.09 & 92.87 & 76.65 & 93.06  
             & 70.64 & 89.55 & 70.92 & 90.00
             & 49.54 & 74.11 & 46.39 & 72.65 \\
    Ours & \textbf{84.94} & \textbf{95.77} & \textbf{85.23} & \textbf{95.76}  
             & \textbf{78.51} & \textbf{92.49} & \textbf{79.35} & \textbf{93.03}
             & \textbf{62.42} & \textbf{83.49} & \textbf{60.15} & \textbf{82.44} \\ \hline \hline
    \textbf{RGB-based} & & & & & & & & & & & &   \\
    I3D~\cite{joze2018ms}  & - & - & 81.76 & 95.16  
            & - & - & 81.97 & 93.79
            & - & - & 57.69 & 81.05 \\
    HMA~\cite{hu2021hand}  & 73.45 & 89.70 & 74.59 & 89.70  
    & 66.30 & 84.03 & 67.47 & 84.03
    & 49.16 & 69.75 & 46.27 & 68.60 \\
    TCK~\cite{li2020transfer}  & 83.04 & 93.46 & 83.91 & 93.52  
            & 80.31 & 91.82 & 81.14 & 92.24
            & - & - & - & - \\ 
    BSL~\cite{albanie2020bsl}  & - & - & - & -  
            & - & - & - & -
            & 64.71 & 85.59 & 61.55 & 84.43  \\
    SignBERT~(+ R)~\cite{hu2021signbert}   & {89.56} & {97.36} & {89.96} & {97.51}
                 & {86.98} & {96.39} & {87.62} & {96.43}
                 & {71.24} & {89.12} & {67.96} & {88.40} \\
    {Ours~(+ R)}~\cite{hu2021signbert}   & \textbf{90.75} & \textbf{97.75} & \textbf{91.52} & \textbf{97.73}
                 & \textbf{88.08} & \textbf{96.47} & \textbf{88.62} & \textbf{96.47}
                 & \textbf{73.71} & \textbf{90.12} & \textbf{70.77} & \textbf{89.30} \\ \hline
    \end{tabular}
    \end{threeparttable}
    \end{center}
    \label{tab:msasl}
    \end{table*}

    \begin{table*}
      \small
      \tabcolsep=6.2pt
      \caption{Evaluation of isolated SLR on WLASL dataset~(the higher the better). I3D~\cite{li2020word} denotes the RGB baseline for fusion.}
      \begin{center}
      \begin{threeparttable}
      \begin{tabular}{l|cc|cc|cc|cc|cc|cc}
      \hline
      \multirow{3}{*}{Method} & \multicolumn{4}{c|}{WLASL100}
                              & \multicolumn{4}{c|}{WLASL300}
                              & \multicolumn{4}{c}{WLASL}\\ \cline{2-13}
              & \multicolumn{2}{c|}{Per-instance} & \multicolumn{2}{c|}{Per-class} 
              & \multicolumn{2}{c|}{Per-instance} & \multicolumn{2}{c|}{Per-class}
              & \multicolumn{2}{c|}{Per-instance} & \multicolumn{2}{c}{Per-class} \\
              & Top-1 & Top-5 & Top-1 & Top-5 
              & Top-1 & Top-5 & Top-1 & Top-5   
              & Top-1 & Top-5 & Top-1 & Top-5    \\ \hline \hline
      \textbf{Pose-based} & & & & & & & & & & & & \\
      ST-GCN~\cite{yan2018spatial} & 50.78 & 79.07 & 51.62 & 79.47   
              & 44.46 & 73.05 & 45.29 & 73.16
              & 34.40 & 66.57 & 32.53 & 65.45 \\ 
      Pose-TGCN~\cite{li2020word} & 55.43 & 78.68 & - & -   
              & 38.32 & 67.51 & - & -
              & 23.65 & 51.75 & - & - \\ 
      PSLR~\cite{tunga2020pose} & 60.15 & 83.98 & - & -   
              & 42.18 & 71.71 & - & -
              & - & - & - & - \\
      SignBERT~\cite{hu2021signbert} & 76.36 & 91.09 & 77.68 & 91.67  
               & 62.72 & 85.18 & 63.43 & 85.71 
               & 39.40 & 73.35 & 36.74 & 72.38 \\
      Ours & \textbf{79.84} & \textbf{92.64} & \textbf{80.72} & \textbf{93.08}  
               & \textbf{73.20} & \textbf{90.42} & \textbf{73.77} & \textbf{90.58} 
               & \textbf{48.85} & \textbf{82.48} & \textbf{46.37} & \textbf{81.33} \\ \hline \hline
      \textbf{RGB-based} & & & & & & & & & & & & \\
      I3D~\cite{li2020word}  & 65.89 & 84.11 & 67.01 & 84.58  
              & 56.14 & 79.94 & 56.24 & 78.38
              & 32.48 & 57.31 & - & - \\
      HMA~\cite{hu2021hand} & - & - & - & -  
               & - & - & - & -
               & 37.91 & 71.26 & 35.90 & 70.00 \\ 
      TCK~\cite{li2020transfer} & 77.52 & 91.08 & 77.55 & 91.42   
              & 68.56 & 89.52 & 68.75 & 89.41
              & - & - & - & - \\ 
      BSL~\cite{albanie2020bsl}  & - & - & - & -  
              & - & - & - & -
              & 46.82 & 79.36 & 44.72 & 78.47 \\
      SignBERT~(+ R)~\cite{hu2021signbert}   & {82.56} & {94.96} & {83.30} & {95.00}
                   & {74.40} & {91.32} & {75.27} & {91.72}
                   & {54.69} & {87.49} & {52.08} & {86.93} \\
      Ours~(+ R)   & \textbf{84.11} & \textbf{96.51} & \textbf{85.05} & \textbf{96.83}
                   & \textbf{78.44} & \textbf{94.31} & \textbf{79.12} & \textbf{94.43}
                   & \textbf{55.59} & \textbf{89.37} & \textbf{53.33} & \textbf{88.82} \\ \hline 
      \end{tabular}
      \end{threeparttable}
      \end{center}
      \label{tab:wlasl}
      \end{table*}

\subsection{Comparison with State-of-the-art Methods}
In this section, we compare our method with previous state-of-the-art methods on three main downstream tasks, including isolated SLR, continuous SLR and SLT.
For comparison, we group them into pose-based and RGB-based methods.

\begin{table}[!t]
  \small
  \caption{Evaluation of isolated SLR on SLR500 dataset~(the higher the better). \cite{hu2020global} denotes the RGB baseline for fusion. }
  \begin{center}
  \tabcolsep=32pt
  \begin{threeparttable}
  \begin{tabular}{l|c}
  \hline
  Method  &  Accuracy   \\  \hline \hline 
  \textbf{Pose-based} \\
  ST-GCN~\cite{yan2018spatial} &  90.0 \\
  SignBERT~\cite{hu2021signbert}      & 94.5     \\ 
  Ours      & \textbf{95.4}     \\ \hline \hline
  \textbf{RGB-based}  \\
  STIP~\cite{laptev2005space}   &  61.8 \\
  GMM-HMM~\cite{tang2015real} &  56.3 \\
  3D-R50~\cite{qiu2017learning}  &  95.1 \\
  HMA~\cite{hu2021hand}   & 95.9     \\ 
  GLE-Net~\cite{hu2020global}   & 96.8      \\ 
  SignBERT~(+ R)~\cite{hu2021signbert}  & {97.7}      \\ 
  Ours~(+ R)  & \textbf{97.8}      \\ \hline 
  \end{tabular}
  \end{threeparttable}
  \end{center}
  \label{tab:slr500}
  \vspace{-0.3cm}
  \end{table}

\subsubsection{Isolated Sign Language Recognition}
\noindent \textbf{Evaluation on MSASL~\cite{joze2018ms}.}
MSASL introduces new challenges given its unconstrained recording conditions.
As illustrated in Table~\ref{tab:msasl}, the accuracy of previous pose-based methods lags largely behind the RGB-based counterpart.
It is mainly attributed to the pose detection failure caused by partially occluded upper body, motion blur and complex background, \emph{etc.}
TCK~\cite{li2020transfer} and BSL~\cite{albanie2020bsl} propose different pre-training techniques on the I3D backbone by leveraging external sign data.
Our method achieves new state-of-the-art performance under both pose-based and RGB-based comparison settings with a notable gain.
Notably, when compared with previous SignBERT~\cite{hu2021signbert}, our framework outperforms it by 12.88\% per-instance Top-1 accuracy improvement on the full set.

\begin{table*}[!t]
  \small
  \caption{Evaluation of SLT on RWTH-PhoenixT dataset (the higher the better). \cite{zhou2021improving} denotes the RGB baseline method for fusion.}
  \vspace{-1.6em}
  \begin{center}
  \tabcolsep=5.8pt
  \begin{threeparttable}
  \begin{tabular}{l|ccccc|ccccc}
  \hline
  \multirow{2}{*}{Method} & \multicolumn{5}{c|}{Dev}
                          & \multicolumn{5}{c}{Test} \\ \cline{2-11}
          & ROUGE & BLEU-1 & BLEU-2 & BLEU-3 & BLEU-4 
          & ROUGE & BLEU-1 & BLEU-2 & BLEU-3 & BLEU-4  \\ \hline \hline
  \textbf{Pose-based} &  &  &  &  &  &  &  &  \\
  Skeletor~\cite{jiang2021skeletor} & 32.66 & 31.97 & 19.53 & 14.01 & 10.91 & 31.80 & 31.86 & 19.11 & 13.49 & 10.35 \\
  Ours & \textbf{45.53} & \textbf{44.45} & \textbf{31.88} & \textbf{24.59} & \textbf{19.86} & \textbf{44.89} & \textbf{44.35} & \textbf{32.09} & \textbf{24.92} & \textbf{20.41} \\ \hline \hline
  \textbf{RGB-based} &  &  &  &  &  &  &  &  \\
  Sign2Text~\cite{cihan2018neural} & 31.80 & 31.87 & 19.11 & 13.16 & 9.94 & 31.80 & 32.24 & 19.03 & 12.83 & 9.58 \\
  TSPNet~\cite{li2020tspnet} & - & - & - & - & - & 34.96 & 36.10 & 23.12 & 16.88 & 13.41 \\
  MCT~\cite{camgoz2020multi} & 45.90 & - & - & - & 19.51 & 43.57 & - & - & - & 18.51 \\
  SL-Trans~\cite{camgoz2020sign} & - & 47.26 & 34.40 & 27.05 & 22.38 & - & 46.61 & 33.73 & 26.19 & 21.32 \\
  BN-TIN-Trans~\cite{zhou2021improving} & 46.87 & 46.90 & 33.98 & 26.49 & 21.78 & 46.98 & 47.57 & 34.64 & 26.78 & 21.68 \\
  SimulSLT~\cite{yin2021simulslt} & 36.04 & 36.01 & 22.60 & 16.05 & 12.39 & 35.13 & 35.92 & 22.70 & 16.03 & 12.27 \\
  PiSLTRc-T~\cite{yin2021simulslt} & 47.89 & 46.51 & 33.78 & 26.78 & 21.48 & 48.13 & 46.22 & 33.56 & 26.04 & 21.29 \\
  STMC~\cite{zhou2021spatial} & 48.24 & 47.60 & 36.43 & 29.18 & 24.08 & 46.65 & 46.98 & 36.09 & 28.70 & 23.65 \\
  SignBT~\cite{zhou2021improving} & 50.29 & 51.11 & 37.90 & 29.80 & 24.45 & 49.54 & 50.80 & 37.75 & 29.72 & 24.32 \\
  Ours~(+ R) & \textbf{51.12} & \textbf{51.46} & \textbf{38.28} & \textbf{30.30} & \textbf{24.95} & \textbf{50.63} & \textbf{52.01} & \textbf{39.19} & \textbf{31.06} & \textbf{25.70} \\ \hline
  \end{tabular}
  \end{threeparttable}
  \end{center}
  \label{tab:phoenixT2}
  \vspace{-1em}
  \end{table*}

\begin{table}[!t]
  \small
  \caption{Evaluation of continuous SLR on RWTH-Phoenix dataset (the lower the better).
  \cite{cui2019deep} denotes the RGB baseline for fusion.}
  \begin{center}
  \tabcolsep=3.5pt
  \begin{tabular}{l|rc|rc}
  \hline
  \multirow{2}{*}{Methods} & \multicolumn{2}{c|}{Dev} & \multicolumn{2}{c}{Test} \\ 
                           &  del / ins   & WER       & del / ins      & WER      \\ \hline\hline
  \textbf{Pose-based} &  &  &  & \\
  P-BLSTM~\cite{cui2019deep}    & 13.4 / 3.5 & 39.6  & 12.3 / 3.4  & 39.3  \\
  P-Trans~\cite{camgoz2020sign} & 16.0 / 3.2 & 40.9  & 14.9 / 3.4  & 40.4  \\
  Ours & 9.0 / 6.3 & \textbf{34.0}  & 7.9 / 6.0  & \textbf{34.1}  \\ \hline \hline
  \textbf{RGB-based} &  &  &  & \\
  CMLLR~\cite{koller2015continuous}  & 21.8 / 3.9 & 55.0  & 20.3 / 4.5      & 53.0  \\
  1-Million-Hand~\cite{koller2016deep-cvpr}  & 16.3 / 4.6 & 47.1  & 15.2 / 4.6      & 45.1    \\
  CNN-Hybrid~\cite{koller2016deep} & 12.6 / 5.1 & 38.3  & 11.1 / 5.7      & 38.8    \\
  SubUNets~\cite{camgoz2017subunets} & 14.6 / 4.0 & 40.8  & 14.3 / 4.0      & 40.7    \\
  RCNN~\cite{cui2017recurrent}  & 13.7 / 7.3 & 39.4 & 12.2 / 7.5 & 38.7 \\
  Re-Sign~\cite{koller2017re}   & -  & 27.1  & -     & 26.8    \\
  Hybrid CNN-HMM~\cite{koller2018deep} & - &31.6 & - & 32.5 \\ 
  CNN-LSTM-HMM~\cite{koller2019weakly}   & - & 26.0  & -  & 26.0    \\
  CTF~\cite{wang2018connectionist}  & 12.8 / 5.2 & 37.9  & 11.9 / 5.6  & 37.8    \\
  Dilated~\cite{pu2018dilated}   & 8.3 / 4.8 & 38.0  & 7.6 / 4.8  & 37.3    \\
  IAN~\cite{pu2019iterative}  & 12.9 / 2.6 & 37.1 & 13.0 / 2.5 & 36.7 \\
  DNF~\cite{cui2019deep}   & 7.8 / 3.5 & 23.8  & 7.8 / 3.4 & 24.4  \\
  FCN~\cite{cheng2020fully} & - & 23.7  & - & 23.9    \\
  CMA~\cite{pu2020boosting}     & 7.3 / 2.7 & 21.3  & 7.3 / 2.4 & 21.9    \\
  PiSLTRc-R~\cite{pan2021pisltrc} & 8.1 / 3.4 & 23.4 & 7.6 / 3.3 & 23.2    \\
  STMC~\cite{zhou2021spatial} & 7.7 / 2.4 & 21.7  & 7.4 / 2.6 & 20.7    \\
  VAC~\cite{min2021visual}    & 7.9 / 2.5 & 21.2  & 8.4 / 2.6 & 22.3    \\
  Ours~(+ R) & 4.8 / 3.7 & \textbf{19.9}  & 4.2 / 3.8 & \textbf{20.0}    \\ \hline
  \end{tabular}
  \end{center}
  \label{tab:phoenix}
  \vspace{-0.2cm}
  \end{table}

  \begin{table}[!t]
    \small
    \caption{Evaluation of continuous SLR on RWTH-PhoenixT dataset (the lower the better).
    \cite{cui2019deep} denotes the RGB baseline for fusion.}
    \begin{center}
    \tabcolsep=7.5pt
    \begin{tabular}{l|rc|rc}
    \hline
    \multirow{2}{*}{Methods} & \multicolumn{2}{c|}{Dev} & \multicolumn{2}{c}{Test} \\ 
                             &  del / ins   & WER       & del / ins      & WER      \\ \hline\hline
    \textbf{Pose-based} &  &  &  & \\
    P-BLSTM~\cite{cui2019deep}    & 13.8 / 3.3 & 40.2 & 12.9 / 3.1  & 40.2  \\
    P-Trans~\cite{camgoz2020sign} & 12.9 / 3.7 & 39.4 & 11.4 / 3.8 & 39.8  \\
    Ours & 9.2 / 4.9 & \textbf{32.9}  & 8.4 / 5.3  & \textbf{33.6}  \\ \hline \hline
    \textbf{RGB-based} &  &  &  & \\
    1-stream~\cite{koller2019weakly}   & - & 24.5  & -  & 26.5    \\
    3-stream~\cite{koller2019weakly}   & - & 22.1  & -  & 24.1    \\
    DNF~\cite{cui2019deep} & 10.5 / 1.9 & 22.7 &  9.8 / 2.4  & 23.5 \\
    SL-Trans~\cite{camgoz2020sign} & 11.7 / 6.5 & 24.9  &  11.2 / 6.1  & 24.6  \\
    FCN~\cite{cheng2020fully}   & - & 23.3  & -  & 25.1    \\
    PiSLTRc-R~\cite{pan2021pisltrc}  & 4.9 / 4.2 & 21.8  & 5.1 / 4.4 & 22.9    \\
    STMC~\cite{zhou2021spatial} & - & 19.6  & -  & 21.0    \\
    Ours~(+ R) & 4.8 / 3.3 & \textbf{18.8}  & 4.3 / 3.9 & \textbf{19.9}    \\ \hline
    \end{tabular}
    \end{center}
    \label{tab:phoenixT1}
    \end{table}

\noindent \textbf{Evaluation on WLASL~\cite{li2020word}.}
WLASL is also the unconstrained recording setting.
Compared with MSASL, it is more challenging with fewer samples and doubled vocabulary size.
As shown in Table~\ref{tab:wlasl}, it is worth mentioning that our single pose-based method even outperforms the most challenging RGB-based method~\cite{li2020transfer, albanie2020bsl}, with over 2\% Top-1 per-instance accuracy improvement on all sets.
When fused with the RGB baseline, the performance of our method further gets improved.

\noindent \textbf{Evaluation on SLR500~\cite{huang2018attention}.}
As demonstrated in Table~\ref{tab:slr500}, STIP~\cite{laptev2005space} and GMM-HMM~\cite{tang2015real} are the traditional methods based on hand-crafted features.
Since this dataset is recorded under the controlled setting, the performance is quite saturated with only Top-1 accuracy reported.
GLE-Net~\cite{hu2020global} is a challenging method, which performs feature enhancement from the global and local views.
Our method achieves 97.8\% Top-1 accuracy, which is new state-of-the-art performance.

In summary, our method greatly shrinks the performance gap between pose-based and RGB-based methods on isolated SLR.
Under the challenging in-the-wild conditions, our method even outperforms the challenging RGB-based methods.
It can be attributed to our designed masking modeling strategies and incorporated prior during pre-training.

\subsubsection{Continuous Sign Language Recognition}
\noindent \textbf{Evaluation on RWTH-Phoenix~\cite{koller2015continuous}.}
As demonstrated in Table~\ref{tab:phoenix}, we exhibit experiment results on RWTH-Phoenix dataset.
Due to the lack of pose-based methods, we adopt two representative RGB-based methods~\cite{cui2019deep,camgoz2020sign} by only changing its visual encoder with the GCN to process pose modality, denoted as ``P-BLSTM'' and ``P-Trans''.
Pose-based methods lag largely behind RGB-based methods, which is largely caused by pose failure caused by the low-quality data and motion blur.
Among pose-based methods, our method largely outperforms the most challenging competitor P-BLSTM with 5.6\% and 5.2\% WER improvement on the dev and test set, respectively.
When fused with the RGB baseline, our method achieves new state-of-the-art performance, \emph{i.e.,} 19.9\% and 20.0\% WER on the dev and test set, respectively.

\noindent \textbf{Evaluation on RWTH-PhoenixT~\cite{cihan2018neural}.}
We make comparison on RWTH-PhoenixT in Table~\ref{tab:phoenixT1}.
This dataset additionally provides spoken German translation corresponding to the sign gloss annotation.
\cite{koller2019weakly} utilizes the spoken translation to infer the mouth shape label, which provides auxiliary cues to recognition.
Besides, it releases multi-stream versions for further performance improvement.
STMC~\cite{zhou2021spatial} also leverages the multi-cue information from the full frame, hand, face and pose and becomes the most challenging competitor.
Our method~(Ours~(+ R)) outperforms it while only utilizing the full video and pose information.

\subsubsection{Sign Language Translation}
\noindent \textbf{Evaluation on RWTH-PhoenixT~\cite{cihan2018neural}.}
As shown in Table~\ref{tab:phoenixT2}, we perform comparison on RWTH-PhoenixT dataset, which is the current most popular benchmark for evaluating SLT.
{Contemporaneous with ours, Skeletor~\cite{jiang2021skeletor} is an influential work that conducts BERT style pre-training but in another field of pose estimation.
Specifically, it inflates the detected poses to 3D ones and conducts BERT-style pre-training with the aim of refining 3D poses.
Then it validates its effectiveness on downstream SLT with the refined poses as input.}
Compared with this challenging pose-based method, our framework directly models the SL statics in the latent semantics space and surpasses it with a larger performance gain, \emph{i.e.,} 8.95\% and 10.06\% BLEU-4 improvement on the dev and test set, respectively.
When compared with RGB-based methods, Ours~(+ R) also achieves new state-of-the-art performance, achieving 24.95\% and 25.70\% BLEU-4 on the dev and test set, respectively.

\subsection{{Evaluation with Deaf Community}}
{The end goal of automatic sign language understanding is to make the daily life of the deaf community more convenient.
Xu~\emph{et al.}~\cite{xu2022automatic} make the first attempt of user study to evaluate the built sign gloss dictionary for sign language learners.
Evaluation participation of the deaf community is also crucial to better analyze our method and outline future work.
In our work, the effectiveness of pre-training is evaluated on the downstream tasks and it is also desirable to perform more direct evaluation on pre-training.}

{To this end, we conduct a user study with the Institutional Review Board~(IRB) approval from our college with granted number No.202200603.
This user study aims to analyze the robustness of our framework under different input noise levels via evaluating the semantic preservation of the pre-training framework output~(pose sequence).
Different noise levels are achieved via choosing different portions of the input tokens to add noise.
There are 10 deaf volunteers participating in this study.
In the study, each volunteer is asked to judge whether the corresponding sign gloss can be identified via observing the framework output.
We report the correct rate to indicate semantic preservation.
Totally, 100 real-world sign videos are collected and involved in this study.
For each video, there are 8 samples, \emph{i.e.,} 4~(noise levels) $\times$ 2~(input and output) needed to be evaluated.}

{As shown in Table~\ref{tab:user_study}, it can be observed that the semantics of the output are well-preserved when the input noise intensity increases, which validates the robustness of our framework.
Meanwhile, the correct rate of the output is consistently better than that of the input under all noise levels.
It reveals that the modeled statistics via our pre-training can bring positive gains on the semantics.
Besides, these deaf participants also give us feedback on the reasons of failed recognition from their perspectives, \emph{e.g.} pose jittering, nonstandard gesture, \emph{etc.}
These results can give us some hints on further improving the pre-training design, \emph{e.g.} inserting the basic gesture types of sign language as the constraint.}

\begin{table}[!t]
  \small
  \tabcolsep=23pt
  \caption{{User study with deaf community on robustness of our pre-training model under different input noise levels. The rate represents the average correct rate which means the deaf volunteer is able to correctly identify the semantic meaning via observing the output pose of our framework.}}
  \begin{center}
  \begin{tabular}{c|cc}
  \hline
  \multicolumn{1}{c|}{\multirow{2}{*}{Noise Intensity}} & \multicolumn{2}{c}{Correct Rate} \\
  & Input & Output   \\ \hline \hline
  0.1 & 23\% & 92\% \\
  0.2 & 16\% & 85\% \\
  0.3 & 10\% & 79\% \\
  0.4 & 4\% & 75\% \\ \hline
  \end{tabular}
  \end{center}
  \label{tab:user_study}
\end{table}

\subsection{Analysis \& Future Work}
The core of our work is modeling the statistics in the sign language domain via maximizing the likelihood of the joint probability distribution, which benefits the downstream sign language understanding tasks.
Despite the success of BERT in NLP, it is non-trivial to leverage its masked modeling pretext task into sign language understanding due to different characteristics between these two domains.
Among them, the major one is information density~\cite{he2021masked}.
Originally, the languages are highly semantic and well represented with 1D sequences of text words, which are defined with clarified semantics.
In contrast, sign pose, expressed in 3D continuous coordinates, is a kind of well-structured data with both spatial and temporal redundancy.
Besides, this kind of signal usually contains noise due to failure estimation.
This fundamental difference raises the following issues to resolve, which outlines potential future work.

\begin{itemize}
  \item Token embedding \& Position encoding.
  These embeddings are needed to be carefully designed considering the hand pose characteristics in the sign language domain, \emph{e.g.} how to effectively represent spatial-temporal positions of hands.
  \item Masking strategy.
  It aims to capture the hierarchical context in the sign data, which needs to consider the characteristic of sign pose data. 
  \item Decoder design \& Pre-training objective.
  The decoder in the pre-training stage performs mapping from the latent feature back to the input.
  In NLP, the decoder predicts the masked discrete words with the cross-entropy objective.
  While for this task, the goal is to reconstruct the continuous sign hand pose sequence. 
  The involved pre-training objective and decoder are needed to design.
\end{itemize}

{\noindent \textbf{More discussion.}
In this work, we provide our solution to the above issues and validate the effectiveness of our framework.
Pre-training on pose has its pros and cons.
Pose data is a semantic and compact representation, which is robust to appearance or background changes and brings potential computation efficiency.
On the other hand, our adopted pose input is estimated by the off-the-shelf pose detector.
Although our framework embeds the capability to capture the cues from the corrupted input pose sequence, its bottleneck is somewhat limited by the quality of the detected pose.
Jointly optimizing the pose detector with our framework may be a possible solution.
It is also desirable to extend masked-modeling-based self-supervised pre-training to RGB data.
Besides, pre-training can go beyond self-supervised learning, \emph{e.g.} multilingual information may be merged as an auxiliary indicator for better performance on downstream tasks. 
}

\section{Conclusion}
In this paper, we propose the \emph{first} self-supervised pretrainable framework with hand prior incorporated, namely SignBERT+.
Given the dominant role of hand during sign language, we take both hands as visual tokens and carefully embed each visual tokens with gesture state and spatial-temporal position information.
Our framework first performs pre-training on a large volume of sign data via reconstructing the masked tokens from the corrupted input sequence.
Specifically, we subtly design hierarchical masked modeling strategies~(joint, frame and clip).
These strategies explicitly consider hand pose characteristics to capture multi-level contextual information.
Furthermore, we design the hand-model-aware decoder to incorporate prior for better optimization and context modeling.
Then, the pre-trained SignBERT+ is fine-tuned for downstream tasks.
Given the task diversities, we design simple yet effective prediction heads on top of the SignBERT+ encoder during fine-tuning.
Extensive experiments are conducted among three main video-based sign language understanding tasks, \emph{i.e,} isolated SLR, continuous SLR and SLT.
Our experiment results demonstrate the effectiveness of our method, achieving new state-of-the-art performance with a notable gain.

\noindent \textbf{Broader Impact.}
It is estimated by World Health Organization~(WHO) that by 2050 over 700 million people will have disabling hearing loss, which accounts for 10\% of global population~\cite{whodeaf}.
The community with hearing loss may feel isolated, lonely and other mental issues when they face the communication barrier in daily life.
One way to assist them is to bridge this gap via the automatic sign language understanding technique.
Our framework is able to promote its development.
However, our technique is not intended for the potential privacy issue, such as surveillance on sign language communication.


%




\ifCLASSOPTIONcaptionsoff
  \newpage
\fi



%



{\small
\bibliographystyle{IEEEtran}
\bibliography{egbib}
}

\end{document}